\ifdefined\pdfminorversion\pdfminorversion=7\fi
\documentclass{article}
\usepackage{iclr2027_conference,times}
\usepackage[utf8]{inputenc}
\usepackage[T1]{fontenc}
\usepackage{microtype}
\usepackage{graphicx}
\usepackage{subcaption}
\usepackage{booktabs}
\usepackage{hyperref}
\hypersetup{hidelinks}
\usepackage{url}
\usepackage{amsfonts}
\usepackage{amsmath}
\usepackage{amssymb}
\usepackage{mathtools}
\usepackage{nicefrac}
\usepackage{xcolor}
\usepackage{algorithm}
\usepackage{algorithmic}
\usepackage{amsthm}
\usepackage{multirow}
\usepackage{placeins}

\newtheorem{proposition}{Proposition}
\newcommand{\adj}{\mathbf{A}}
\newcommand{\calF}{\mathcal{F}}
\newcommand{\calA}{\mathcal{A}}
\newcommand{\calG}{\mathcal{G}}
\newcommand{\calI}{\mathcal{I}}
\newcommand{\calV}{\mathcal{V}}
\newcommand{\TypeA}{\textsc{Type-A}}
\newcommand{\TypeB}{\textsc{Type-B}}
\newcommand{\Kseeds}{5}
\newcommand{\Pwidth}{16}
\newcommand{\Bwidth}{16}
\newcommand{\Ddepth}{3}
\newcommand{\ConsensusWeight}{16}
\newcommand{\MaxPolicyEvals}{\number\numexpr\Kseeds*(1+\Bwidth*(\Ddepth-1))\relax}
\newcommand{\MaxRetainedStates}{\number\numexpr\Kseeds*(1+\Bwidth*\Ddepth)\relax}
\newcommand{\NplibHitOne}{29.64}
\newcommand{\NplibHitTen}{50.81}
\newcommand{\NplibReach}{57.78}
\newcommand{\NplibFPHitOne}{28.02}
\newcommand{\MsgHitOne}{30.22}
\newcommand{\MsgHitTen}{41.33}
\newcommand{\MsgReach}{46.88}
\newcommand{\MsgFPHitOne}{29.47}
\newcommand{\NplibDirectHitOne}{26.03}
\newcommand{\NplibDirectHitTen}{49.07}
\newcommand{\NplibDirectReach}{54.17}
\newcommand{\NplibDirectPipelineSeconds}{2,524.59}
\newcommand{\NplibDirectPipelineRatio}{12.62}
\newcommand{\NplibDirectPipelineRatioRounded}{12.6}
\newcommand{\NplibSmallPipelineSeconds}{81.11}
\newcommand{\MsgDirectPipelineRatio}{21.36}
\newcommand{\MsgDirectPipelineRatioRounded}{21.4}
\newcommand{\MsgSmallPipelineSeconds}{1,456.71}
\newcommand{\NplibMatchedWidthPipelineRatio}{31.1}
\newcommand{\MsgMatchedWidthPipelineRatio}{60.0}
\newcommand{\PipelineRatioRange}{13--21}
\newcommand{\NplibFrigidHitOneGain}{4.61}
\newcommand{\NplibFrigidHitTenGain}{17.44}
\newcommand{\NplibIcebergNativeHitOne}{24.03}
\newcommand{\NplibIcebergNativeHitTen}{33.13}
\newcommand{\NplibFrigidZeroSeedMisses}{45}
\newcommand{\NplibFrigidInitialHitOne}{20.42}
\newcommand{\NplibFrigidInitialHitTen}{23.91}
\newcommand{\NplibFrigidRefinedHitOne}{28.02}
\newcommand{\NplibFrigidRefinedHitTen}{43.34}
\newcommand{\NplibIcebergZeroSeedMisses}{23}
\newcommand{\NplibIcebergInitialHitOne}{22.54}
\newcommand{\NplibIcebergInitialHitTen}{30.14}
\newcommand{\NplibIcebergRefinedHitOne}{28.27}
\newcommand{\NplibIcebergRefinedHitTen}{46.82}
\newcommand{\NplibMsgptRefinedHitOne}{26.28}
\newcommand{\MsgMsgptRefinedHitOne}{25.56}
\newcommand{\NplibPipelineSeconds}{200.12}
\newcommand{\NplibPipelinePerSpectrum}{0.249}
\newcommand{\NplibPipelineMinutes}{3.3}
\newcommand{\MsgPipelineSeconds}{4,091.00}
\newcommand{\MsgPipelinePerSpectrum}{0.233}
\newcommand{\NplibFrigidBaseSeconds}{5,353}
\newcommand{\NplibFrigidWallSeconds}{94,408}
\newcommand{\NplibFrigidWallHours}{26.2}
\newcommand{\UntrainedHitOne}{\ensuremath{4.61\pm2.90}}
\newcommand{\UntrainedReach}{\ensuremath{8.30\pm5.29}}

\title{GEMS: Molecular Structure Elucidation as Learned Search over Isomer Graphs}

\iclrfinalcopy
\author{Utku Umur Acikalin\textsuperscript{1} \quad
Yingheng Wang\textsuperscript{1} \quad
Goncalo J. Gouveia\textsuperscript{2,3} \\
\bfseries Tyler J. Schwertfeger\textsuperscript{2,3} \quad
Frank C. Schroeder\textsuperscript{2,3} \quad
Carla P. Gomes\textsuperscript{1} \\
\normalfont\textsuperscript{1}Department of Computer Science, Cornell University \\
\normalfont\textsuperscript{2}Department of Chemistry and Chemical Biology, Cornell University \\
\normalfont\textsuperscript{3}Boyce Thompson Institute, Cornell University}
\hypersetup{
  pdftitle={GEMS: Molecular Structure Elucidation as Learned Search over Isomer Graphs},
  pdfauthor={Utku Umur Acikalin, Yingheng Wang, Goncalo J. Gouveia, Tyler J. Schwertfeger, Frank C. Schroeder, Carla P. Gomes}
}

\begin{document}

\maketitle
\lhead{Preprint}
\vspace{-2.5\baselineskip}
\raggedbottom

\begin{abstract}
Identifying small-molecule structures from molecular formulas and tandem mass
spectra remains a bottleneck in untargeted metabolomics and natural-product
discovery. Even when database retrieval or de novo prediction does not identify
the exact molecular structure, the resulting candidates can provide starting
points for exploring nearby structures through a few bond edits.
We introduce GEMS (Guided Editing of Molecular Structures), which refines
candidates through learned 2-switch bond edits that preserve the molecular
formula and each atom's total valence. Molecular fingerprints encode local
structural features and can be computed directly from candidate structures.
GEMS uses MIST, a pretrained spectrum-to-fingerprint model, to predict fingerprint
probabilities once per spectrum. A neural policy compares each candidate's fingerprint
with these predictions, links present features to the atoms and bonds that
generate them, and scores all valid edits in one forward pass. The policy is
trained on shortest edit paths between structures, so it can be pretrained on
PubChem isomers without spectra and then fine-tuned with spectrum-predicted
fingerprints. At inference,
GEMS searches from multiple initial candidates matching the supplied molecular
formula and ranks candidates by fingerprint agreement and multi-start consensus.
With the target structure excluded from PubChem initialization, three edit
rounds achieve Hit@1/Hit@10 of \NplibHitOne\%/\NplibHitTen\% on NPLIB1 and
\MsgHitOne\%/\MsgHitTen\% on MassSpecGym, exceeding published baselines under
comparable settings. On NPLIB1, GEMS achieves higher identification accuracy
in about \NplibPipelineMinutes{} minutes end to end, compared with
\NplibFrigidWallHours{} hours for our FRIGID run. GEMS also improves identification from
generated candidates, including those refined by FRIGID. Experiments with exact target fingerprints
identify fingerprint prediction error as a bottleneck in target recovery and
final ranking. Deeper search recovers additional targets without retraining,
with larger gains from weaker initial candidates.
\par\vspace{-0.5\baselineskip}
\end{abstract}

\section{Introduction}
\label{sec:intro}

Identifying unknown small molecules is a central challenge in untargeted
metabolomics and natural-product discovery. Many measured spectra remain
unassigned to structures, limiting what experiments reveal about a sample's
chemistry \citep{bushuiev2024massspecgym}. Tandem mass spectrometry (MS/MS)
measures the mass-to-charge ratios and intensities of fragment ions, providing
clues about substructures whose arrangement can remain ambiguous. We study
the known-formula setting, which fixes elemental composition while leaving
connectivity to be determined. Even then, many isomers can be consistent
with the observed spectrum.

Database retrieval ranks existing structures, so success requires the
target to be in the database
\citep{duhrkop2015sirius,goldman2023mist}. Generative methods can propose new
structures, but recovering exact connectivity remains difficult
\citep{stravs2022msnovelist,diffms,msgpt2026}. Yet even when these methods miss the target, their candidates are often only a few
bond edits away. For each NPLIB1 spectrum, we remove the target from
PubChem and retrieve up to five isomers whose fingerprints best match the one
predicted from the spectrum.
For at least 72.85\% of spectra, one of these isomers lies within three bond edits of the target (Appendix~\ref{app:seed_proximity_evidence}). 
FRIGID, a recent de novo method
\citep{frigid2026}, shows a similar pattern in its initial candidate pool:
among spectra with no correct selected starting candidate, at least 47.79\%
have a starting candidate within three bond edits of the target
(Appendix~\ref{app:generator_refinement}).

These near misses suggest correcting candidates locally rather than discarding
them. Two recent methods do this with ICEBERG, a structure-to-spectrum model \citep{goldman2024iceberg}.
FRIGID remasks and regenerates candidate regions \citep{frigid2026}, and FOAM
mutates candidates and selects them by predicted spectra
\citep{manjrekar2026foam}. We instead supply candidate--fingerprint
correspondences directly to a learned policy, which learns to select bond edits
from this evidence without simulating spectra.

\begin{figure}[t]
  \centering
  \includegraphics[width=\linewidth]{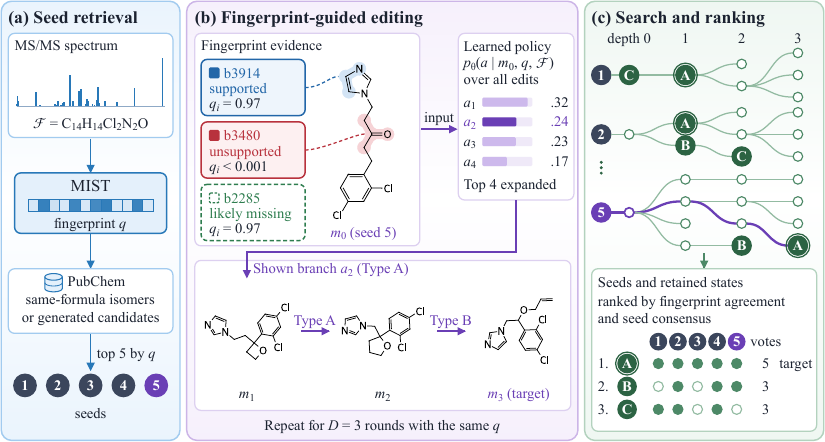}
\caption{\textbf{GEMS: fingerprint-guided search over molecular isomers.}
(a) MIST predicts fingerprint probabilities $q$ once from the MS$^2$ spectrum
and molecular formula $\calF$. Fingerprint likelihood selects five same-formula
seeds, excluding the target.
(b) Supported present bits have $x_i=1$ and high $q_i$, unsupported present
bits have $x_i=1$ and low $q_i$, and likely missing bits have $x_i=0$ and
high $q_i$. Present bits are mapped to the atoms and bonds that generate them,
while missing bits have no local occurrence. One policy evaluation scores all
valid 2-switch edits. For Seed~5, the four highest-ranked edits receive
96.4\% of policy probability, and each begins a shortest three-edit path to the
reference structure. The illustrated path (purple) starts
with $a_2$.
(c) Initial and retained states are pooled and
ranked by $s(m;q)+16\log(1+c(m))$, where $s(m;q)$ measures fingerprint agreement
and $c(m)$ counts the distinct seeds whose searches reach $m$, including
through their initial states. The target ranks first among 33 distinct
candidate structures. Panel (c) omits Seeds~3 and~4, which still vote. NPLIB1 query
\texttt{CCMSLIB00001058451}. The example uses $K=5$ starts, proposal and beam
widths $P=B=4$, and depth $D=3$. The main experiments use $P=B=16$.}
  \label{fig:overview}
\end{figure}

We introduce GEMS (Guided Editing of Molecular Structures), which formulates
structure elucidation as learned search over molecular isomers
(Figure~\ref{fig:overview}). Each search step applies a 2-switch edit, which
removes one bond unit from each of two bonds and reconnects their endpoints. One
bond may be to a hydrogen. These edits preserve the molecular formula
and each atom's total valence. MIST, a pretrained spectrum-to-fingerprint model,
predicts the target fingerprint probabilities once per spectrum
\citep{goldman2023mist}. The policy compares these probabilities with the
candidate's fingerprint and links the candidate's present features to their generating atoms
and bonds. A present feature with low predicted probability,
for example, points to where an edit may help. A candidate can admit hundreds or
thousands of valid edits, all scored in one forward pass without
fingerprinting every possible product. GEMS runs beam searches from
several initial candidates, then ranks the retained molecules by fingerprint
agreement and by how many independent searches reach them.

We train the policy on shortest edit paths between molecular structures. Several
edits can be equally good first steps toward the target, so rather than picking
one as the label, we maximize the total probability of all edits that begin a
shortest path. Both this supervision and the exact target fingerprint can be
computed from structures alone, so we pretrain the policy on pairs of PubChem
isomers without any spectra, then fine-tune it on benchmark training spectra with
MIST-predicted fingerprints. Because the best next edit depends only on the
current molecule and the target, the policy is Markovian: its inputs, including
the fingerprint correspondence, are computed from the current molecule and
predicted fingerprint, with no trajectory history or depth index. The same policy
can therefore search at greater depths and refine retrieved or generated
candidates without retraining or changing the generator architecture.

Our contributions are: \textbf{(1)} We formulate known-formula structure
elucidation as search over an isomer graph whose edges are formula- and
valence-preserving 2-switch edits, allowing retrieved or generated candidates
to be corrected locally. Because one close start is enough, we search from
several starts, and even a simple Enumerate \& Score (E\&S) search, which
scores every valid one-edit product by fingerprint agreement without a learned
policy, matches or exceeds the compared published baselines on the same splits
in Hit@1 and Hit@10. \textbf{(2)} We introduce GEMS, whose policy
links present fingerprint features to the atoms and bonds that generate them
and scores all valid edits in one forward pass. Even from PubChem starts that exclude every isomer
with Tanimoto similarity of at least 0.7 to the target, GEMS exceeds
all Table~\ref{tab:main_results} baselines in Hit@1 and Hit@10 on both datasets.
It ranks more targets first than E\&S at matched width, and outperforms it at
wider width with \PipelineRatioRange{}-fold lower runtime.
From FRIGID's initial NPLIB1 candidates, without database retrieval, GEMS also
outperforms FRIGID's refinement of the same pool. \textbf{(3)} We introduce a pretraining strategy that uses all shortest edit paths
between pairs of structures as supervision, so it requires no spectra. A policy
pretrained only on PubChem isomers, without training on any spectra, already
exceeds all Table~\ref{tab:main_results} baselines on NPLIB1 in Hit@1 and Hit@10
(28.89\% Hit@1). When both are trained from scratch on experimental NPLIB1
spectra, this supervision also outperforms REINFORCE with exact-fingerprint rewards. \textbf{(4)} We study how start
quality, search width and depth, and fingerprint accuracy shape recovery and
identification. Deeper search helps most from weak starts but raises recovery more than
identification, while exact fingerprints improve both.

\section{Related Work}
\label{sec:related}

\paragraph{Candidate ranking.}
CSI:FingerID and MIST predict molecular fingerprints from MS/MS spectra to
rank database candidates
\citep{duhrkop2015sirius,goldman2023mist}. Other approaches, including CFM-ID,
ICEBERG, MassFormer, FraGNNet, GLACIER, and AIMe, use spectrum prediction to
support ranking by spectral similarity
\citep{allen2015cfmid,goldman2024iceberg,young2024massformer,young2024fragnnet,
wang2026glacier,acikalin2026aime}. These approaches rank a supplied candidate
set. GEMS uses MIST retrieval to select starting structures, then edits their
connectivity to explore additional isomers.

\paragraph{De novo generation.}
MSNovelist, Spec2Mol, and MS2Mol generate molecular strings from spectra or
predicted fingerprints \citep{stravs2022msnovelist,spec2mol,ms2mol}.
DiffMS and MBGen generate molecular graphs through diffusion, while FlowMS
uses discrete flow matching \citep{diffms,mbgen,flowms}. All three use MIST
encoders and pretrain their decoders on fingerprint--structure pairs.
FRIGID-base (without ICEBERG refinement) and MS-GPT generate SAFE strings,
a fragment-based molecular representation, from predicted fingerprints through
masked diffusion and autoregressive decoding, respectively
\citep{frigid2026,msgpt2026}. MS-GPT handles fingerprint uncertainty
through multiple calibrated binary queries. Generated structures can
initialize GEMS, which explores additional isomers through learned bond edits.

\paragraph{Scaffold conditioning and candidate refinement.}
MADGEN completes a retrieved scaffold, while MSAnchor predicts a scaffold and
its attachment sites to guide generation \citep{madgen,msanchor2026}.
FRIGID refines complete candidates using ICEBERG fragment predictions
\citep{frigid2026,goldman2024iceberg}. It uses discrepancies between a
candidate's predicted fragments and the experimental spectrum to set masking
probabilities through a fixed rule, and its learned diffusion model then
regenerates the masked regions from the predicted fingerprint and formula,
without explicitly receiving the discrepancy information. FOAM searches from
retrieved or generated structures through random formula-preserving mutations
and crossovers, and uses ICEBERG-predicted spectra, together with a
structural-complexity penalty, only to select candidates \citep{manjrekar2026foam}.
GEMS learns to select valid bond edits from candidate--fingerprint
correspondences, using the same MIST fingerprint prediction throughout search,
without simulating candidate spectra.

\paragraph{Molecular graph editing.}
Chemical graph enumeration has a long history, from DENDRAL to Surge
\citep{lindsay1993dendral,mckay2022surge}, and learned graph edits have also been
used for chemical reaction prediction \citep{do2019graph2graph}.
GEMS builds on classical degree-preserving 2-switches
\citep{senior1951partitions,hakimi1962realizability,
        taylor1980constrained,fernandes2022switch}, adapting them to molecular
bonds and attached hydrogens and learning which valid edits to apply from
spectral evidence.

\section{Structure Elucidation as Search over Isomers}
\label{sec:formulation}

\subsection{Molecular Candidates and Bond Edits}
\label{sec:actions}

We represent each candidate by its heavy atoms, integer bond orders, and each heavy atom's hydrogen count. We consider connected,
neutral molecules without isotopic labels or radicals, and evaluate
connectivity with stereochemistry removed. Each search step applies a
formula-preserving bond edit.

\begin{figure}[htbp]
  \centering
  \includegraphics[width=\linewidth]{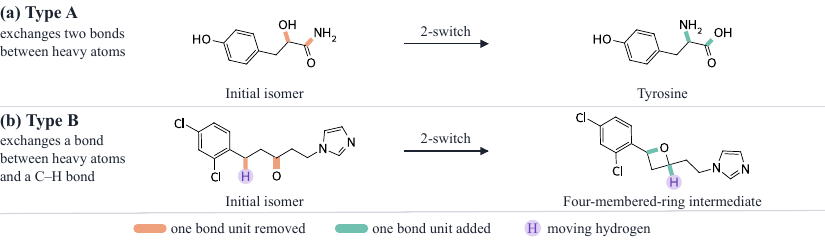}
\caption{\textbf{Formula- and valence-preserving 2-switch edits.}
(a) \TypeA{} exchanges hydroxyl and amino attachments to reach tyrosine. (b) \TypeB{} exchanges one carbonyl bond unit and one C--H bond for a new C--O
bond and a new C--H bond, forming a four-membered ring. This is action $a_1$
from Seed~5 in Figure~\ref{fig:overview}.
Each affected atom loses and gains one bond unit. Orange marks removed bond units,
teal marks added ones, and purple marks the moving hydrogen.}
  \label{fig:2switch}
\end{figure}

We use two types of 2-switch edits (Figure~\ref{fig:2switch}), where a bond
unit is one bond-order unit. \TypeA{} removes one bond unit from each
of two heavy-atom bonds, $(u,v)$ and $(w,z)$, with four distinct atoms, and
adds one unit each to $(u,w)$ and $(v,z)$. The other pairing adds $(u,z)$
and $(v,w)$ instead.

\TypeB{} uses the same switch with a hydrogen endpoint. Let
$u$, $v$, and $w$ be distinct heavy atoms, with a hydrogen on $u$
and a bond between $v$ and $w$. The edit removes the $u$--H bond, reduces the
$v$--$w$ bond order by one, adds a $u$--$v$ bond unit, and attaches
the hydrogen to $w$. Exchanging $v$ and $w$ gives the other pairing.
For example, keto--enol interconversion can be represented by one
\TypeB{} edit.

In both edit types, every affected atom loses and gains one bond unit.
The edits therefore preserve the formula and each atom's total bond order,
including attached hydrogens. We retain products with nonnegative integer
bond orders up to three, nonnegative attached-hydrogen counts, and a connected
heavy-atom graph. Appendix~\ref{app:state_representation} gives the matrix
representation and invariants.

\subsection{Search over Isomers}
\label{sec:properties}
\label{def:isomer_graph}

Applying a valid edit produces another candidate with the same molecular
formula. These transitions define an implicit isomer graph
$\calG_{\calF}=(\calV_{\calF},\mathcal E_{\calF})$. Each node is a molecular
state as defined in Section~\ref{sec:actions} with formula $\calF$, and an
edge corresponds to a valid edit between two states. A sequence of edits
traces a path through this graph. The edits also preserve each atom's total
valence, so sharing a molecular formula does not guarantee that two states
can be connected by a path of valid edits.

We enumerate outgoing edits only when expanding a state. For $E$ heavy-atom
bond units and $n$ heavy atoms, there are $O(E^2)$ potential \TypeA{} and
$O(nE)$ potential \TypeB{} edits, each checked for validity. Rapid neighborhood
growth (Appendix~\ref{app:neighborhood_example}) motivates learned beam search
that selects edits before fingerprinting their products.

\subsection{Candidate Recovery and Ranking}
\label{sec:problem}

Search starts from initial isomers and returns a pool of candidate structures.
It \emph{reaches} the target when the pool contains a structure with the target's
InChIKey14, the first 14 characters of the InChIKey. GEMS can fail at three
stages. Initialization may supply no candidate from which the target is
reachable within the search depth. Search may discard a path to the target.
Ranking may place other candidates above a recovered target. We examine
initialization and depth in Section~\ref{sec:seed_proximity}, and fingerprint
errors in navigation and ranking in Section~\ref{sec:ablations}.

\section{GEMS}
\label{sec:method}

At each step, the candidate molecule specifies present local structural
features and their locations. Fingerprint probabilities predicted from the
experimental spectrum indicate the unknown target's likely features. The navigation policy uses their agreement and
discrepancies to choose bond edits.

Given a spectrum $S$ and molecular formula $\calF$, GEMS uses MIST, a frozen
spectrum-to-fingerprint model, to predict $q\in[0,1]^{4096}$. These probabilities
are computed once and remain fixed throughout search. GEMS selects initial
isomers, searches from each independently, and ranks the retained candidates
by fingerprint agreement and support from distinct starts:
\begin{equation}
\begin{aligned}
  q &= g_{\mathrm{MIST}}(S,\calF), \\
  \mathcal M_0 &= \calI(q,\calF), \\
  \mathcal C_\theta
    &= \operatorname{Search}_\theta(q,\mathcal M_0,\calG_{\calF}), \\
  \widehat m
    &= \arg\max_{m\in\mathcal C_\theta} R(m;q).
\end{aligned}
\label{eq:decomposition}
\end{equation}

\subsection{Initialization}
\label{sec:initialization}

GEMS starts from chemically valid molecules with formula $\calF$. These can
come from database retrieval or a de novo generator such as
FRIGID~\citep{frigid2026} or MS-GPT~\citep{msgpt2026}. Our main configuration
selects initial PubChem isomers by agreement with the predicted
fingerprint.

For a candidate fingerprint $x(m)\in\{0,1\}^{4096}$, we measure agreement with
$q$ using the independent-bit Bernoulli log likelihood
\begin{equation}
    s(m;q)=\sum_i \bigl[x_i(m)\log\widetilde q_i
      +(1-x_i(m))\log(1-\widetilde q_i)\bigr],
    \label{eq:bernoulli}
\end{equation}
where $\widetilde q_i=\operatorname{clip}(q_i,10^{-7},1-10^{-7})$.
This score is used for both retrieval and final ranking.

We rank PubChem molecules with formula $\calF$ by $s(m;q)$, retain the
highest-scoring representative of each InChIKey14 group, and select up to
$K=\Kseeds{}$ initial structures. In target-excluded evaluation, we remove
the target InChIKey14 group before selecting the starts. A correct prediction
therefore requires at least one edit to a retrieved structure.

\subsection{Candidate--Fingerprint Correspondence}
\label{sec:correspondence}

GEMS compares the candidate's radius-two, 4,096-bit Morgan fingerprint $x(m)$
\citep{rogers2010ecfp} with the spectrum-predicted probabilities $q$
(Figure~\ref{fig:overview}b). \emph{Supported present bits} occur in the
candidate and have high $q_i$, while \emph{unsupported present bits} occur in
the candidate and have low $q_i$. \emph{Likely missing bits} are absent from
the candidate but have high $q_i$. Together, they suggest which features to
preserve, change, or introduce.

An atom environment comprises a center atom and surrounding atoms and
bonds within zero, one, or two bond steps. For each present fingerprint bit,
we retain all environments that generate it, linking the bit to the
candidate's atoms and bonds. In Figure~\ref{fig:overview}(b), bit 3480 is
present in Seed~5 and corresponds to a radius-one environment centered on
its carbonyl carbon, but has predicted probability below 0.001. The policy
uses this localized evidence to score edits. Missing bits have no corresponding
environment in the candidate, but still provide target evidence through
their fingerprint tokens.

The policy receives 128 fingerprint tokens, initially allocated as 48 likely
missing, 48 unsupported present, and 32 supported present bits
(Appendix~\ref{app:correspondence}). Each encodes bit identity, role, $q_i$,
$x_i$, and uncertainty and discrepancy features.

\subsection{Fingerprint-Conditioned Navigation Policy}
\label{sec:policy}

A transformer encodes the fingerprint tokens, and a Graphormer-style molecular
encoder \citep{ying2021graphormer} processes the candidate. Atom representations
attend to fingerprint tokens with learned biases distinguishing environment
centers, other members, and atoms outside the environment. Directed-bond
representations are then constructed from atom representations and bond
features. They attend to fingerprint tokens using separate biases that
describe whether a bond and its endpoints belong to an environment. These
connections let the policy associate predicted fingerprint support with the
atoms and bonds involved in an edit. Appendix~\ref{app:network} gives the
layer arrangement.

Each valid edit is scored from the representations of the atoms and bonds it
changes. We concatenate the relevant representations with action-type and
global molecular features and combine them using multilayer perceptrons.
The same edit can be written with its endpoints in different orders. To make
its score independent of these equivalent orderings, we average their action
representations before projecting to a logit. A softmax over the valid edit
set $\calA(m)$ gives
\begin{equation}
  p_\theta(a\mid m,q,\calF)
  =\operatorname{softmax}_{a\in\calA(m)}
      \ell_\theta(a;m,q,\calF).
  \label{eq:policy}
\end{equation}
One forward pass scores all valid edits. Product molecules are constructed
and fingerprinted only for actions selected by search. The Markovian
policy conditions on the current molecule, fingerprint probabilities, and
formula, without trajectory history or a depth index.

\subsection{Training the Navigation Policy}
\label{sec:training}

During training, the chosen target graph $t$ is known. We identify edits that
move a source molecule $m$ one step closer to $t$ along a shortest valid path.
Several edits may be equally good: if three edits are required, every first
edit that leaves a two-edit path to $t$ is optimal. Let $\calA^+(m;t)$ contain
all such optimal first edits. We maximize their total probability by minimizing
\begin{equation}
    \mathcal L_{\mathrm{nav}}
    =-\log\sum_{a\in\calA^+(m;t)}
       p_\theta(a\mid m,q,\calF).
    \label{eq:loss}
\end{equation}
This objective does not require choosing a single optimal edit as the
label. In Figure~\ref{fig:overview}(b), Seed~5's four highest-ranked edits each
begin a shortest three-edit path and jointly receive 96.4\% of policy
probability. D$k$ denotes sources whose shortest valid path to the chosen
target graph $t$ has $k$ edits.\footnote{Training paths exclude held-out
structures and match the chosen target graph up to atom renumbering.
Evaluation accepts any structure with the target's InChIKey14, so evaluation
distances can be shorter (Appendix~\ref{app:target_equivalence}).}

Structure-only pretraining first teaches the policy
to navigate with exact target fingerprints. Fine-tuning then adapts it to
fingerprints predicted by a frozen MIST encoder from experimental spectra.

\paragraph{PubChem pretraining.}
We mine PubChem isomer pairs one or two edits apart and keep all optimal first
edits for each. The input is the exact target fingerprint, $q=x(t)$, so no
spectra are needed. If each source had a single target, the policy
could learn a molecule's typical edits and ignore the fingerprint.
Half the examples therefore come in groups that share a source but have
different targets, preferably with different optimal edits, encouraging the
policy to use the target fingerprint. The other half selects one pair per
sampled source group to cover more formulas and molecules. Both benchmark
models start from one shared checkpoint, selected on validation after
37.5 million examples (Appendix~\ref{app:training}).

\paragraph{Fine-tuning with predicted fingerprints.}
Fine-tuning pairs benchmark training targets with same-formula PubChem sources
one to three edits away, using fingerprint probabilities predicted from the
target's spectrum by the frozen MIST encoder. It varies the source for a fixed
target, complementing the fixed-source, varying-target groups in pretraining.
We find shortest paths by meeting in the middle: the target is expanded by
up to two edits, each source by one, and the neighborhoods are matched.
Targets above 35 heavy atoms use D1 and D2 only, because two-step expansion is
costly. We keep all available D1 and D2 observations and a capped number of D3
sources per spectrum, selected by MIST Bernoulli agreement to approximate how
starting candidates are selected at inference (Appendix~\ref{app:supervision}).

\subsection{Multi-Start Policy Beam Search}
\label{sec:search}

GEMS runs an independent beam search from each of $K$ initial structures,
with proposal width $P$, beam width $B$, and maximum depth $D$.
Our main setting uses $K=\Kseeds{}$, $P=\Pwidth{}$, $B=\Bwidth{}$, and
$D=\Ddepth{}$.
For a trajectory $\pi_d=(m_0,a_1,m_1,\ldots,a_d,m_d)$, the score is the
cumulative policy log probability:
\begin{equation}
  L_\theta(\pi_d;q,\calF)
  =\sum_{t=1}^{d}
    \log p_\theta(a_t\mid m_{t-1},q,\calF).
  \label{eq:path_policy}
\end{equation}
At each depth, the policy scores all valid edits from each beam state.
We construct product molecules and compute their fingerprints only for the
top $P$ selected actions. Within each start's search, previously visited
products are removed, and duplicate products retain their highest-scoring
trajectory. The top $B$ trajectories under $L_\theta$ form the next beam.
The visited set records all proposed products, including those outside the
next beam (Algorithm~\ref{alg:search}, Appendix~\ref{app:search}).

The final candidate pool contains the initial structures and beam states
retained at every depth across all starts. With our main setting, search
uses at most $K[1+B(D-1)]=\MaxPolicyEvals{}$ state-level policy evaluations
and retains at most $K(1+BD)=\MaxRetainedStates{}$ states before final
deduplication.

\subsection{Final Ranking by Fingerprint Evidence and Consensus}
\label{sec:ranking}

Searches from different seeds can reach the same molecule. This
agreement is an additional ranking signal. Let $c(m)$ count the distinct
seeds whose retained search pools contain $m$, including the seeds themselves.
Each seed contributes at most one vote. In Figure~\ref{fig:overview}(c),
candidate A receives five votes and candidate C receives three, including
a vote because C is Seed~1's initial structure. Support merges atom renumberings
and alternative Kekul\'e assignments while keeping tautomers distinct
(Appendix~\ref{app:path_matching}).

The final score is $R(m;q)=s(m;q)+\lambda\log(1+c(m))$,
where $s(m;q)$ measures fingerprint agreement and $\lambda$ weights consensus.
We use $\lambda=\ConsensusWeight{}$ for both benchmarks, selected on MassSpecGym validation. We retain each InChIKey14 group's highest-scoring candidate
and rank these representatives by $R(m;q)$.

\section{Experiments}
\label{sec:experiments}

\subsection{Setup and Evaluation}
\label{sec:setup}

\paragraph{Datasets and metrics.}
We use the same NPLIB1 split as FRIGID, with 803 protonated test spectra
\citep{duhrkop2021canopus,frigid2026}, and the structurally held-out MassSpecGym
split with all 17,556 test spectra, 14,066 protonated and 3,490 sodiated
\citep{bushuiev2024massspecgym}. One MassSpecGym policy serves both adducts.
Hit@$k$ measures correctness among the first $k$ ranked connectivities.
All queries contribute to accuracy and reach, including 17 NPLIB1 and 200
MassSpecGym queries without starts. MassSpecGym hit counts are pooled across
adducts.

\paragraph{Comparisons.}
Published baselines come from FRIGID Table~1 and \emph{MassSpecGym in the Wild}
Table~9 \citep{frigid2026,liu2026massspecgymwild}. Appendix~\ref{app:structural_metrics}
reports MS-GPT's published results and our best-effort reproduction
with corrected masking and retraining on the same benchmark splits.

\subsection{Main Results}
\label{sec:main_results}

GEMS achieves the highest Hit@1 and Hit@10 in Table~\ref{tab:main_results} on both benchmarks. Relative to the
strongest baseline, FRIGID on NPLIB1 and our MS-GPT reproduction on MassSpecGym, the gains
are \NplibFrigidHitOneGain{}/\NplibFrigidHitTenGain{} and 6.99/13.63 percentage points. Because PubChem starts exclude the target
connectivity, every hit requires at least one edit.

\begin{table}[t]
  \caption{Known-formula structure elucidation (H@$k$ in \%). GEMS rows use $K=5$, $P=B=16$, $D=3$,
and $\lambda=16$. PubChem starts exclude the target connectivity, and Tani.\ $<0.7$ also
excludes starts with target similarity $\geq 0.7$ (Appendix~\ref{app:filtered_starts}). Generator $\to$ GEMS
starts from MIST-selected outputs, which may contain the target. $\dagger$: our same-split
reproduction. Bold marks the best per dataset. Lower maximum common edge subgraph
(MCES) distance and higher Tanimoto similarity are better. Empty outputs receive MCES 100.
Baselines are from \citet{liu2026massspecgymwild}
and \citet{frigid2026}.}
  \label{tab:main_results}
  \centering
  \footnotesize
  \renewcommand{\arraystretch}{0.9}
  \setlength{\tabcolsep}{4pt}
  \begin{tabular*}{\linewidth}{@{\extracolsep{\fill}}lrrrrrr}
    \toprule
    Method & H@1 & MCES@1 & Tani.@1 & H@10 & MCES@10 & Tani.@10 \\
    \midrule
    \multicolumn{7}{l}{\textbf{NPLIB1}} \\
    MIST + MolForge & 2.24 & 14.16 & 0.40 & 5.11 & 10.96 & 0.47 \\
    DiffMS & 8.34 & 10.46 & 0.35 & 15.44 & 8.54 & 0.47 \\
    FRIGID-base & 19.80 & 7.88 & 0.54 & 24.41 & 6.20 & 0.61 \\
    FRIGID & 25.03 & \textbf{7.10} & 0.58 & 33.37 & \textbf{5.56} & 0.64 \\
    MS-GPT$^{\dagger}$ & 21.05 & 18.49 & 0.50 & 28.64 & 17.79 & 0.54 \\
    \midrule
    MS-GPT$^{\dagger}$ $\rightarrow$ GEMS & 26.28 & 20.61 & 0.52 & 42.22 & 19.54 & 0.60 \\
    FRIGID-base $\rightarrow$ GEMS & 28.02 & 11.38 & 0.56 & 43.34 & 10.23 & 0.64 \\
    GEMS (PubChem, Tani. $<0.7$) & 27.90 & 8.59 & 0.56 & 46.45 & 6.97 & 0.67 \\
    \textbf{GEMS} & \textbf{29.64} & 7.73 & \textbf{0.59} & \textbf{50.81} & 6.08 & \textbf{0.71} \\
    \midrule
    \multicolumn{7}{l}{\textbf{MassSpecGym}} \\
    DiffMS & 2.30 & 13.96 & 0.28 & 4.25 & 11.68 & 0.39 \\
    FOAM (DiffMS + PubChem) & 1.50 & 12.21 & 0.35 & 10.28 & \textbf{6.06} & 0.53 \\
    MIST + MolForge (corrected) & 10.73 & 22.15 & 0.37 & 14.48 & 17.88 & 0.41 \\
    FRIGID-base & 16.09 & 15.02 & 0.41 & 18.19 & 11.95 & 0.46 \\
    FRIGID & 18.29 & 13.49 & 0.43 & 22.00 & 11.65 & 0.47 \\
    MS-GPT$^{\dagger}$ & 23.23 & 25.15 & 0.48 & 27.70 & 24.71 & 0.50 \\
    \midrule
    MS-GPT$^{\dagger}$ $\rightarrow$ GEMS & 25.56 & 26.66 & 0.48 & 34.07 & 25.91 & 0.52 \\
    GEMS (PubChem, Tani. $<0.7$) & 23.93 & 10.95 & 0.50 & 31.75 & 9.61 & 0.55 \\
    \textbf{GEMS} & \textbf{30.22} & \textbf{9.68} & \textbf{0.55} & \textbf{41.33} & 8.26 & \textbf{0.61} \\
    \bottomrule
  \end{tabular*}
\end{table}

GEMS also has the highest Top-10 Tanimoto similarity among these methods:
0.71 versus 0.64 for FRIGID on NPLIB1, and 0.61 versus 0.53 (FOAM) on MassSpecGym.
MCES results are mixed across benchmarks and ranks, with full comparisons
in Appendix~\ref{app:structural_metrics}.

With starts of Tanimoto similarity $\geq 0.7$ to the target also excluded,
GEMS still exceeds every baseline in Table~\ref{tab:main_results} in Hit@1
and Hit@10 on both benchmarks (Appendix~\ref{app:filtered_starts}).

\subsection{Refining Generator Initializations}
\label{sec:generator_refinement}

GEMS refines generated candidates without retraining
(Table~\ref{tab:generator_refinement}). From the NPLIB1 FRIGID-base
pool, it raises Hit@1/Hit@10 of the selected starts from
\NplibFrigidInitialHitOne\%/\NplibFrigidInitialHitTen\% to
\NplibFrigidRefinedHitOne\%/\NplibFrigidRefinedHitTen\%, above
\NplibIcebergNativeHitOne\%/\NplibIcebergNativeHitTen\% for our FRIGID run
after 24 ICEBERG refinement rounds, and those of FRIGID's final pool from
\NplibIcebergInitialHitOne\%/\NplibIcebergInitialHitTen\% to
\NplibIcebergRefinedHitOne\%/\NplibIcebergRefinedHitTen\%. For our MS-GPT
reproduction, it raises Hit@1 from 21.05\% to \NplibMsgptRefinedHitOne\%
on NPLIB1 and from 23.23\% to \MsgMsgptRefinedHitOne\% on MassSpecGym
(Appendix~\ref{app:generator_refinement}).

\subsection{Learned Proposals Enable Wider Search}
\label{sec:direct_search}

We compare learned proposals with Enumerate \& Score (E\&S) from the same
five MIST-ranked, target-excluded starts. E\&S constructs and fingerprints
every valid one-edit neighbor, retaining the best $B$ trajectories per start
by fingerprint log likelihood. Both methods search three edits deep and rank
retained candidates with $\lambda=16$. We test E\&S widths $B\in\{1,2,4\}$
and GEMS widths $P=B\in\{4,16\}$ (Table~\ref{tab:direct_search}). Every tested
E\&S width exceeds the published baselines in Table~\ref{tab:main_results}
in Hit@1 and Hit@10 on both benchmarks.

At width sixteen, GEMS exceeds all tested E\&S widths in Hit@1, Hit@10, and
reach on both benchmarks. Against E\&S $B=4$, NPLIB1 Hit@1/Hit@10 rises from
\NplibDirectHitOne\%/\NplibDirectHitTen\% to \NplibHitOne\%/\NplibHitTen\%,
and reach from \NplibDirectReach\% to \NplibReach\%. MassSpecGym Hit@1 rises
from 28.16\% to 30.22\% and reach from 44.24\% to 46.88\%.
Learned proposals make the wider beam practical: GEMS takes \NplibPipelineSeconds{} s
on NPLIB1 versus \NplibDirectPipelineSeconds{} s for E\&S ($B=4$), a
\NplibDirectPipelineRatio{}-fold difference. The MassSpecGym ratio is
\MsgDirectPipelineRatio{} (Appendix~\ref{app:runtime}).

Random-weight policies achieve \UntrainedHitOne\% NPLIB1 Hit@1 under the same
proposal and beam budget (Appendix~\ref{sec:pretraining_control}).

\subsection{Initialization, Search Width, and Depth}
\label{sec:seed_proximity}

At depth three, up to five MIST-ranked starts from the same-formula PubChem
pool reach 57.78\% of NPLIB1 targets. Ranking within a random 250-isomer subset
reaches 50.44\%, and using five random starts reaches 22.17\%
(Table~\ref{tab:initialization_depth}). At width sixteen, extending search
from three to ten rounds adds 18.68 points of reach from up to five random starts but
2.12 points from full-pool MIST-ranked starts.

We next vary width and depth from up to five MIST-ranked starts or up to
100 random starts, keeping the checkpoint and each starting set fixed
(Figure~\ref{fig:width_depth}, Appendix~\ref{app:width_depth}). From $P=B=D=3$,
widening to $P=B=10$ at $D=3$ and deepening to $D=10$ at $P=B=3$ each retain
at most 30 search states plus the initial structure per start. Under MIST
guidance, widening gives higher Hit@1 than deepening from ranked starts
(29.89\% versus 29.02\%), while deepening gives higher Hit@1 from up to 100
random starts (29.39\% versus 25.90\%). With exact fingerprints, deepening
helps more from both starting sets. Hit@10 and per-seed reach follow the same
ordering. The relative benefits of width and depth therefore depend on
initialization and fingerprint evidence.

With up to five MIST-ranked starts, D1--D2/D2--D3 adds 2.37/0.25 Hit@1
points on NPLIB1 and 0.34/$-0.21$ on MassSpecGym. Removing starts within
one edit of the target makes D1 Hit@1 zero by construction; D2/D3 reaches
24.16\%/25.53\% and 20.31\%/22.44\%, respectively
(Appendix~\ref{app:filtered_starts}).

\subsection{Fingerprint Errors Limit Search and Ranking}
\label{sec:ablations}

MIST fingerprints guide retrieval, navigation, and ranking. We keep the
MIST-selected starts and trained checkpoint fixed and replace the navigation
fingerprint, ranking fingerprint, or both with the exact target fingerprint.
A ranking-only substitution preserves the retained pool and consensus counts.

\begin{figure}[ht]
  \centering
  \includegraphics[width=\linewidth]{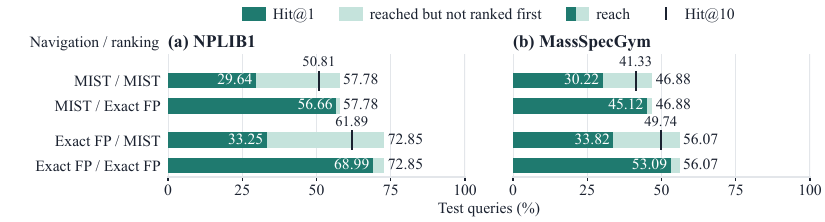}
\caption{\textbf{Finding the target and ranking it first.}
Dark bars show Hit@1, total bar length gives reach, and ticks show
Hit@10 for rows ranked with MIST. Labels specify navigation/ranking fingerprints.
Exact FP is computed from the target as an oracle. Each dataset uses fixed
starts and checkpoint, $K=5$, $P=B=16$, $D=3$, and $\lambda=16$.
All 803 NPLIB1 and 17,556 MassSpecGym queries count, including queries without starts.
Table~\ref{tab:fingerprint_oracle} gives the values.}
  \label{fig:fingerprint_oracle}
\end{figure}

On fixed candidate pools, exact-fingerprint ranking raises Hit@1 from 29.64\% to 56.66\% on
NPLIB1 and from 30.22\% to 45.12\% on MassSpecGym (Figure~\ref{fig:fingerprint_oracle}), placing the target first for
nearly all recovered queries. Exact navigation raises reach from 57.78\% to 72.85\% and from
46.88\% to 56.07\%, respectively. With MIST ranking, however, Hit@1 rises only to 33.25\%
and 33.82\%. Fingerprint prediction error thus limits both recovery and final ranking
(Appendix~\ref{app:fingerprint_ablation}). Reranking the top ten with an ICEBERG spectral score adds 2.24 and 1.38 Hit@1
points on the two benchmarks (Appendix~\ref{app:iceberg_reranking}).

\section{Discussion and Limitations}
\label{sec:discussion}

GEMS shows that local correction of complete candidates is effective for known-formula structure
elucidation. Needing only same-formula structures and a predicted fingerprint, one policy refines
retrieved candidates and outputs of generators with different architectures. It makes wider search
practical and remains useful at greater depths with weak starts. The exact-fingerprint interventions
point to better spectrum-to-fingerprint prediction as a route to better recovery and ranking.

The method assumes a known formula and evaluates connectivity without
stereochemistry. Formula uncertainty would require searching and ranking
across multiple formula hypotheses. The policy reflects its PubChem training
data, PubChem initialization depends on available same-formula candidates,
and valence-preserving edits restrict the states reachable from each start.

\section*{Ethics Statement}

This work develops computational methods for molecular annotation in
metabolomics and natural-product research. More accurate structure proposals
may accelerate discovery, but predictions require experimental validation and
should not be the sole evidence for safety-critical chemical identification.

\section*{AI Use Statement}

Generative AI assisted implementation and debugging, experiment planning and
coordination, analysis, literature search, and drafting and editing manuscript
text. The authors tested the code, checked numerical results against experiment
outputs and bibliographic claims against primary sources, and take
responsibility for the final content.

\section*{Reproducibility Statement}

Appendices~\ref{app:state_details}--\ref{app:search} specify molecular states,
supervision construction, model inputs, training, and search.
Appendices~\ref{app:supporting_results}--\ref{app:runtime} give full results,
evaluation conventions, and timing boundaries.
Upon publication, we will publicly release our code, trained GEMS checkpoints,
and supporting artifacts, including experiment configurations, starting-candidate
lists, and data-preparation scripts, to enable reproduction of the reported results.

\FloatBarrier
\bibliography{references}
\bibliographystyle{iclr2027_conference}

\clearpage
\appendix

\section{Molecular States and Target Equivalence}
\label{app:state_details}

\subsection{Molecular Representation and Transition Invariants}
\label{app:state_representation}

A molecular state contains an ordered tuple of heavy-atom symbols and an
integer adjacency matrix $\adj$. For $i\neq j$, $A_{ij}\in\{0,1,2,3\}$ gives
the bond multiplicity between heavy atoms $i$ and $j$. The diagonal entry
$A_{ii}$ gives the number of hydrogens attached to atom $i$. Atom indices
specify the endpoints affected by an individual edit. Every retained state is
connected, neutral, non-isotopic, and non-radical. Stereochemistry is removed.

\begin{proposition}[Transition invariants]
\label{prop:invariants}
For every retained transition $m'=T(m,a)$ and aligned atom $i$,
\begin{equation}
  \mathsf{form}(m')=\mathsf{form}(m),
  \qquad
  A'_{ii}+\sum_{j\neq i}A'_{ij}
  =A_{ii}+\sum_{j\neq i}A_{ij}.
  \label{eq:transition_invariants}
\end{equation}
Thus each action preserves the molecular formula and the total bond order
incident to every aligned atom.
\end{proposition}
The equalities follow from the balanced bond-unit subtraction and addition in
both action types. The transition filter separately enforces supported bond
orders and connectedness.

\subsection{Training and Evaluation Target Equivalence}
\label{app:target_equivalence}

Let $\mathcal T_{\mathrm{train}}(t)$ contain the states isomorphic to the chosen
target graph, preserving heavy-atom symbols, integer bond orders, and attached
hydrogens. Atom numbering may differ, but a match must preserve the complete
literal graph. The matching key preserves the chosen Kekul\'e assignment and tautomer.
The evaluation target class is
\begin{equation}
  \mathcal T_{\mathrm{eval}}(t)
  =\{u\in\calV_{\calF}:\operatorname{IK14}(u)=\operatorname{IK14}(t)\}.
\end{equation}
The training class is contained in the evaluation class.

Write $\calG_{\calF}^{\mathrm{train}}$ for the edit graph restricted to states
admitted by the training exclusions, including removal of held-out
intermediates. For admitted sources and targets, define
\begin{equation}
\begin{aligned}
 d_{\mathrm{train}}(s,t)
   &=\min_{u\in\mathcal T_{\mathrm{train}}(t)}
     d_{\calG_{\calF}^{\mathrm{train}}}(s,u),\\
 d_{\mathrm{eval}}(s,t)
   &=\min_{u\in\mathcal T_{\mathrm{eval}}(t)}
     d_{\calG_{\calF}}(s,u).
\end{aligned}
\end{equation}
An unreachable target class has infinite distance. Every admissible training
path is also an admissible evaluation path, and the evaluation goal is broader.
Consequently,
\begin{equation}
 d_{\mathrm{eval}}(s,t)\leq d_{\mathrm{train}}(s,t).
 \label{eq:distance_goal_relation}
\end{equation}
A source can require three edits to reach the chosen training target and only
two to reach another evaluation-equivalent structure. 

\subsection{Path Matching and Search Deduplication}
\label{app:path_matching}

Training joins use a canonical key for the literal integer-bond graph without
aromatizing or changing its Kekul\'e assignment. A proposed match must admit an
atom mapping that preserves the complete adjacency matrix and heavy-atom
symbols. Target-side actions are reversed and transported through that mapping
into the source coordinates. Each resulting forward action is checked against
the valid edit set, intermediate exclusions are enforced, and the final state
must match the reference target graph.

Online beam search and consensus use canonical stereo-free aromatic SMILES as
a structural key. This key identifies alternative Kekul\'e assignments and
keeps tautomers distinct. Search retains one concrete
state for each selected key and applies subsequent edits to that state.
Retaining one representation per key can discard distinct edit continuations.
InChIKey14 is used for target exclusion and connectivity evaluation.
Search paths consist of valid edits between concrete molecular states.

\begin{table}[htbp]
  \caption{Identity conventions. Training exclusions additionally restrict
  admissible intermediate states.}
  \label{tab:identity_conventions}
  \centering
  \scriptsize
  \begin{tabular}{p{0.23\linewidth}p{0.40\linewidth}p{0.27\linewidth}}
    \toprule
    Identity & Representations merged & Use \\
    \midrule
    Literal integer-bond graph & Atom renumberings preserving elements,
    bond orders, and attached hydrogens. The chosen Kekul\'e assignment is
    retained. & Training path joins and distance labels \\
    Canonical stereo-free aromatic SMILES & Atom renumberings and alternative
    Kekul\'e assignments. Tautomers remain distinct. & Search deduplication
    and consensus votes \\
    InChIKey14 & Structures assigned the same connectivity block by InChI,
    including some tautomeric forms & Target exclusion, recovery, and final
    connectivity ranking \\
    \bottomrule
  \end{tabular}
\end{table}

\section{Supervision Construction}
\label{app:supervision}

\subsection{Identity, Validity, and Held-Out Exclusion}

Supervision and online search use the same action enumeration, application,
and validation code. Appendix~\ref{app:state_details} defines the molecular
states, matching rules, and identity conventions.

Before PubChem pretraining pairs are mined, every validation and test
connectivity from both evaluation benchmarks is excluded from source and
target structures. Generated intermediates with an excluded connectivity are
also removed. 

\subsection{Exact-Fingerprint PubChem Pairs}

For each formula, every valid one-edit product of every remaining PubChem
structure is generated and identical exact products are merged. Two PubChem
structures connected directly define a directed D1 pair. Two structures whose
expansions meet at a shared product define a D2 pair when no direct transition
exists. All valid first actions that realize a shortest direct connection or
enter a shortest shared intermediate are retained as positives.

Pairs are grouped by source molecule. Half of the sampled examples come from
contrast groups containing up to four D1 or D2 targets for the same source,
prioritizing different sets of optimal first edits and a mixture of distances.
The other half selects one pair from each sampled group to broaden coverage
across formulas and source molecules, preferring a D1 target when present and
otherwise using a D2 target. Each example uses the exact target fingerprint.

The mined corpus contains 986,681,148 directed source--target examples.
Pretraining samples 125 million examples with a 50:50 contrast/coverage
allocation. Contrast groups contribute at most four examples and coverage
groups one.

\subsection{Target-Conditioned D1--D3 Supervision}

A distance shell contains states at a given shortest-path distance from the
target. For a fixed target $t$, define these shells and optimal first-action
sets as
\begin{equation}
\begin{aligned}
  \mathcal D_k(t)
    &= \{m:d_{\mathrm{train}}(m,t)=k\}, \\
  \calA^+(m;t)
    &= \{a\in\calA(m):T(m,a)\in\mathcal D_{k-1}(t)\},
       \quad m\in\mathcal D_k(t),\ k\geq1.
\end{aligned}
\label{eq:target_shells}
\end{equation}
Here $d_{\mathrm{train}}$ uses the target class and training exclusions defined
in Appendix~\ref{app:target_equivalence}. Thus $\mathcal D_0(t)$ contains the
atom-renumbered versions of the reference target graph.

For the D1--D3 corpus, containing molecules with at most 35 heavy atoms,
each chosen training target is expanded through two edits. States are matched by the literal bond-and-hydrogen graph up to atom
renumbering. Held-out intermediates are excluded during construction, so the
resulting shells are complete in the admissible training graph. Reusing these
target neighborhoods avoids expanding every PubChem source through two steps.

Same-formula, target-excluded PubChem sources matching the target's D1 or D2
shell receive the corresponding distance. Every source is expanded once and
its products are matched to the target shells. For a source outside the
two-step neighborhood, a valid join to the D2 shell gives a three-step path
after reversing the target-side path and mapping its atom indices to those
of the source-side meeting state. Shorter shell membership is checked first, so the joined path
certifies distance three to the chosen training goal. Every retained path is
replayed forward under the edit rules and exclusions described in
Appendix~\ref{app:path_matching}.

For a source $m\in\mathcal D_k(t)$, every valid first action entering the
complete $\mathcal D_{k-1}(t)$ shell is positive. All such first actions are
retained for each selected training pair, with a replayed path witnessing each
one. 
\paragraph{Larger molecules.}
For molecules with more than 35 heavy atoms, we construct a separate D1--D2
corpus to avoid the cost of exhaustive two-step target expansion. We enumerate
the one-edit neighbors of each target and each same-formula PubChem source.
A direct source--target connection gives D1 supervision. If no direct connection
exists, a shared one-edit neighbor gives a shortest two-edit path. All source
edits entering such shared neighbors are retained as positive actions, and the
corresponding paths are replayed and verified. Thus, one-step expansion on each
side supplies supervision at distances one and two without a target D2 shell.

\paragraph{Training observations.}
D3 sources are capped independently for each target spectrum by
\begin{equation}
  N_{\mathrm{D3}}^{\mathrm{keep}}
  =\min\!\left(N_{\mathrm{D3}},\max(50,N_{\mathrm{D2}})\right).
  \label{eq:d3_cap}
\end{equation}
The D3 sets contain 378,563 source--target pairs for NPLIB1 and
1,677,732 for MassSpecGym. After pairing these with the target spectra,
the cap retains 155,625 and 6,486,972 training observations, respectively. Membership is selected by descending
MIST Bernoulli agreement for each spectrum, so spectra of the same target can
select different D3 sources. All available D1 and D2 observations are retained.

MIST predicts the target fingerprint from the spectrum, formula, and
formula-based peak annotations. The target structure supplies the shortest-path
labels. Training excludes source--target pairs at distance zero and mixes
D1--D3 examples for molecules with at most 35 heavy atoms with D1--D2 examples
for molecules with more than 35 heavy atoms, using the batch schedules in
Appendix~\ref{app:training}.

\section{Fingerprint Correspondence Details}
\label{app:correspondence}

The candidate fingerprint is a binary radius-two Morgan fingerprint with
4,096 hashed bits, bond types enabled, and chirality disabled
\citep{rogers2010ecfp}. Each of the 128 fingerprint tokens encodes a bit's
identity, evidence type, target probability, candidate presence, and
uncertainty and discrepancy features. We initially allocate 48 tokens to
high-probability bits missing from the candidate, 48 to present bits with
low predicted support, and 32 to present bits with high predicted support.
Unfilled slots are filled from the remaining bits by confidence ranking.

The numerical features for a selected bit are
\[
(q_i, x_i, q_i(1-x_i), (1-q_i)x_i, q_ix_i, q_i-x_i, h(q_i)),
\]
where $h(q)=-q\log q-(1-q)\log(1-q)$ is Bernoulli entropy, with
$h(0)=h(1)=0$. The discrepancy feature is signed.

The occurrence map retains every environment that generates each present bit.
Radius-zero membership contains only the center atom. Radius-one and
radius-two membership expand through the candidate adjacency and include the
corresponding heavy bonds. These relations connect fingerprint evidence to
atom and directed-bond tokens in the policy.
Atom relations distinguish centers and other members at each radius from
atoms outside the environment. Bond relations record how many endpoints
belong to an environment and whether the bond itself belongs. Missing bits
have a separate relation code because they have no occurrence in the candidate.

\section{Training Protocol}
\label{app:training}

\subsection{MIST Encoders and Input Processing}

We use the released FRIGID MIST encoders \texttt{mist\_canopus.pt} for NPLIB1
and \texttt{mist\_msg.pt} for MassSpecGym.
Each benchmark uses one fixed MIST encoder for all training and evaluation
spectra.
NPLIB1 uses public formula-only peak assignments. MassSpecGym training uses
formula-only subformula assignment with 15 ppm tolerance. Protonated evaluation spectra use 20 ppm on both benchmarks.
Sodiated MassSpecGym evaluation uses 15 ppm. Peak assignment and fingerprint
prediction use the spectrum and formula.
Sodiated inference uses the same MassSpecGym encoder with corrected padding
and pairwise formula masks, a single square-root intensity transform, the raw
Top-60 peaks, and all peaks retained after annotation.
Fingerprint prediction uses the sodium adduct. The navigation policy does not
take adduct as an input. Adduct-dependent evidence enters through the MIST
fingerprint probabilities.

\subsection{Network Configuration}
\label{app:network}

All reported fingerprint policies use the same architecture. The embedding
width is 256, the feed-forward width is 512, and every attention block uses
eight heads. The target fingerprint encoder has two self-attention layers. The molecular encoder has six layers before the two atom--fingerprint
cross-attention layers and two layers afterward. Directed bond
representations use two self-attention and two fingerprint cross-attention
layers. Residual and attention dropout are both 0.1. Atom and bond
cross-attention gates and the global target gate are initialized with logit
$-2$. Collision energy is not used.

Action heads concatenate directed-bond representations, action-type
embeddings, and global molecular features before an MLP. Type A averages
four equivalent endpoint orderings. Type B first constructs a hydrogen-transfer
representation from donor and recipient atom states and the global state,
combines it with directed-bond representations, and averages two orderings.
The averaged action representation is projected to a logit.

\subsection{PubChem Pretraining}

Pretraining uses four H100 GPUs with distributed data parallelism, global
batch size 256, and AdamW with learning rate $4\times10^{-4}$, weight decay
$10^{-2}$, and $\beta=(0.9,0.999)$. The learning rate follows cosine decay
to zero, starting at the specified rate. Gradient norm is clipped at 1.
We use full-precision training, one batch per optimizer step, and random seed 42.
The run processes 125 million examples in 488,282 optimizer steps,
saving checkpoints at 20 evenly spaced training milestones.
Downstream training begins from the checkpoint after 37,499,904 examples,
selected by validation Hit@1. We initialize the policy weights
from this checkpoint and reset the optimizer, learning-rate schedule, and
random states before fine-tuning.

\subsection{Target-Conditioned Fine-Tuning}

Fine-tuning minimizes Equation~\ref{eq:loss} and alternates batches from
the D1--D3 set (at most 35 heavy atoms) and the D1--D2 set (more than 35
heavy atoms).
NPLIB1 uses a 63:1 batch ratio, and MassSpecGym uses a 15:1 ratio.
After removing distance-zero pairs, the combined sets contain 314,220
training observations for NPLIB1 and 13,412,519 for MassSpecGym.
NPLIB1 trains for three epochs with 5,746 steps per epoch, and MassSpecGym
trains for one epoch with 221,855 steps. The reported NPLIB1 checkpoint is
from step 8,619, halfway through epoch two. The MassSpecGym checkpoint is
from the end of its training epoch.

Both models use one H100, batch size 64, learning rate $5\times10^{-5}$,
AdamW with weight decay $10^{-2}$ and $\beta=(0.9,0.999)$, and cosine decay
to zero from the initial learning rate. Gradient norm is clipped at 1.
We use full precision, one batch per optimizer step, seed 42, and two data
loader workers with unpinned batches.

\section{Search Algorithm}
\label{app:search}

Valid-action enumeration checks bond-order bounds and heavy-atom connectivity
using temporary adjacency arrays. Search materializes product states and
constructs RDKit molecules for fingerprinting only for selected top-$P$
actions. These proposals can later be removed by duplicate/visited checks or
beam pruning. Trajectories use policy log probability to preserve sequences
of learned edit choices, which need not monotonically improve the current
molecule's fingerprint agreement.

\begin{algorithm}[ht]
\caption{Multi-start policy beam search}
\label{alg:search}
\begin{algorithmic}[1]
\REQUIRE Fingerprint probabilities $q$, formula $\calF$, initial isomers
$\{m_0^k\}_{k=1}^K$, policy $p_\theta$, proposal width $P$, beam width $B$,
depth $D$
\STATE $\mathcal C\leftarrow\{m_0^1,\ldots,m_0^K\}$
\FOR{$k=1,\ldots,K$}
  \STATE $\mathcal H_k\leftarrow\{(m_0^k,0)\}$. $\mathcal V_k\leftarrow\{m_0^k\}$
\ENDFOR
\FOR{$d=1,\ldots,D$}
  \FOR{$k=1,\ldots,K$}
    \STATE $\mathcal P_k\leftarrow\emptyset$
    \FOR{$(m,L)\in\mathcal H_k$}
      \STATE enumerate valid edits $\calA(m)$
      \FOR{top-$P$ actions $a$ under $p_\theta(\cdot\mid m,q,\calF)$}
        \STATE $m'\leftarrow T(m,a)$.
        $L'\leftarrow L+\log p_\theta(a\mid m,q,\calF)$
        \STATE add $(m',L')$ to $\mathcal P_k$
      \ENDFOR
    \ENDFOR
    \STATE remove members of $\mathcal V_k$ and deduplicate $\mathcal P_k$,
    retaining the largest $L'$ per structure
    \STATE $\mathcal H_k\leftarrow$ top-$B$ members of $\mathcal P_k$ by $L'$
    \STATE $\mathcal V_k\leftarrow\mathcal V_k\cup
    \{m':(m',L')\in\mathcal P_k\}$
    \STATE $\mathcal C\leftarrow\mathcal C\cup
    \{m':(m',L')\in\mathcal H_k\}$
  \ENDFOR
\ENDFOR
\STATE compute multi-start support $c(m)$ for every $m\in\mathcal C$,
  including each seed's depth-zero visit
\STATE \textbf{return} unique InChIKey14 classes ranked using the score in
  Section~\ref{sec:ranking}
\end{algorithmic}
\end{algorithm}

\section{Evaluation Protocol and Baseline Comparisons}
\label{app:supporting_results}

\subsection{Conventions}
\label{app:conventions}

Unless stated otherwise, GEMS uses up to $K=5$ MIST-ranked, target-excluded
PubChem starts, $P=B=16$, $D=3$, $\lambda=16$, and the reported checkpoint for
each benchmark. Query-level results use all 803 NPLIB1 and 17,556 MassSpecGym
test queries. MassSpecGym results pool 14,066 protonated and 3,490 sodiated
spectra. Queries without starts count as failures. With PubChem
initialization, these are 17 NPLIB1 and 200 MassSpecGym queries (129
protonated and 71 sodiated). Appendix~\ref{app:generator_refinement} gives the
corresponding counts for generated pools.

Reach is the fraction of queries whose retained candidate pool contains a
structure with the target's InChIKey14. It does not depend on ranking and is an
upper bound on Hit@$k$. Per-seed reach, used in Appendix~\ref{app:depth_analysis},
is the fraction of seed--target pairs whose own retained lineage contains the
target. Queries without starts contribute no pairs, and queries with more
starts receive more weight.

Results at a depth $D$ below the searched maximum are prefixes of longer
searches, with candidates and consensus votes restricted to states retained by
depth $D$. A target first recovered at depth $D$ shows a path of $D$ edits, but
the beam may have discarded a shorter one.

Labels such as MIST and Exact FP name the fingerprint used for navigation,
ranking, or both. Exact FP is computed from the target structure. It is an
oracle diagnostic that is unavailable at inference. Starting sets are fixed
across corresponding MIST and exact-fingerprint conditions. Ranked starting
sets are selected with MIST. Each training setting uses one random seed.
The random-five comparison in Appendix~\ref{app:initialization_sensitivity}
reports repeated starting-set selections.

\subsection{Structural Metrics and Published Baselines}
\label{app:structural_metrics}

Tables~\ref{tab:published_nplib1} and~\ref{tab:published_massspecgym} give the
complete published comparisons. NPLIB1 values come from FRIGID Table~1.
MassSpecGym values come from \emph{MassSpecGym in the Wild} Table~9, with
FRIGID-base from FRIGID Table~1. We use the corrected MolForge results.
\emph{MS-GPT (published)} provides contextual comparisons
\citep{msgpt2026}, with a different NPLIB1 split from ours. For numerical
comparisons and refinement experiments, we use \emph{MS-GPT (ours)}, our
best-effort reproduction retrained on the same benchmark splits after
correcting the MIST attention masking issue described below. This reproduction
has not been validated by the original authors.
The reproduction uses the same full test sets and FRIGID-compatible structural
metric implementation as GEMS. The NPLIB1 adapter checkpoint was selected
on its 396 validation spectra, and the MassSpecGym checkpoint by validation
cross-entropy.

\paragraph{MIST masking in MS-GPT.}
\label{app:mist_implementation}
The released MS-GPT implementation retains the MIST padding-mask issue
documented in \emph{MassSpecGym in the Wild}
\citep[Section~6.1.1]{liu2026massspecgymwild}.\footnote{Code inspected at
\href{https://github.com/VIKI623/MS-GPT/blob/f2c844e61f079051baf3014c33df5ba2f0524363/src/ms_gpt/external/mist/transformer_layer.py\#L630-L635}{MS-GPT commit \texttt{f2c844e}}.}
Padded tokens can contribute to attention, making fingerprint predictions
depend on batch padding. GEMS excludes padded tokens from attention. The effect of the released
masking code on MS-GPT's reported accuracy has not been measured.

GEMS structural metrics use the ranked top-ten unique InChIKey14 candidates
and include every test query. We reuse FRIGID's implementation with RDKit
2026.3.5, \texttt{myopic\_mces} 1.2.2, and PuLP 3.3.2. Molecules are converted
through InChI, tautomer-canonicalized using FRIGID's two-transform canonicalizer,
and stripped of stereochemistry. Tanimoto uses binary radius-two Morgan
fingerprints with 2,048 bits. For each query, Top-$k$ Tanimoto is the maximum
similarity among its first $k$ candidates, averaged over all queries.

MCES (maximum common edge subgraph) distance measures differences in molecular
bonding under the best correspondence between atoms of the same element,
accounting for bond order. Lower values indicate closer structural agreement,
including when a candidate is not an exact match.

We report thresholded MCES with threshold 15. FRIGID's evaluator removes
invalid or disconnected structures and duplicate InChIs from the first ten
ranked candidates, preserving order without backfilling. Thresholded scoring
can return lower bounds instead of exact distances. If canonicalization or
MCES computation fails, the evaluator substitutes a bond-count score: the
sum of target and candidate bond counts, or twice the target bond count.
For each query, Top-$k$ MCES is the minimum returned score among the first
$k$ remaining candidates, capped at 100 and averaged over all test queries. Queries
without eligible candidates receive MCES 100 and Tanimoto zero. In our
best-effort MS-GPT reproduction, 106 NPLIB1 and 3,295 MassSpecGym queries have
no eligible candidates after structural preprocessing. Their MCES-100
penalties contribute 13.20 and 18.77, respectively, to the reported mean MCES.

The full results show the same pattern as Table~\ref{tab:main_results}: GEMS
improves exact-match accuracy and Top-10 Tanimoto over the baselines evaluated
on the same splits. FRIGID has lower MCES on NPLIB1. On MassSpecGym, GEMS reduces both
MCES measures relative to FRIGID, while FOAM has lower Top-10 MCES
(6.06 versus 8.26).

\begin{table}[t]
  \caption{NPLIB1 known-formula baselines and GEMS variants. Non-MS-GPT published values come from FRIGID Table~1 \citep{frigid2026}. $\dagger$: our best-effort same-split MS-GPT reproduction with corrected masking and retraining. $\ddagger$: published MS-GPT values are contextual comparisons with protocol qualifications in this section \citep{msgpt2026}. All GEMS variants use up to five starts, $P=B=16$, $D=3$, and $\lambda=16$. PubChem starts exclude the target connectivity. The one-edit row additionally excludes every target-specific one-edit neighbor connectivity group, without a Tanimoto filter. Generator and Tanimoto-filtered starts follow Table~\ref{tab:main_results}. Our evaluations retain all test queries. Best displayed values are bold, including ties.}
  \label{tab:published_nplib1}
  \centering
  \scriptsize
  \setlength{\tabcolsep}{4pt}
  \begin{tabular}{lrrrrrr}
    \toprule
    Method & H@1 & MCES@1 & Tani.@1 & H@10 & MCES@10 & Tani.@10 \\
    \midrule
    Spec2Mol & 0.00 & 48.57 & 0.12 & 0.00 & 17.58 & 0.16 \\
    MADGEN & 2.10 & 20.56 & 0.22 & 2.39 & 12.69 & 0.27 \\
    MIST + Neuraldecipher & 2.32 & 10.26 & 0.35 & 6.11 & 8.53 & 0.43 \\
    MIST + MolForge & 2.24 & 14.16 & 0.40 & 5.11 & 10.96 & 0.47 \\
    MIST + MSNovelist & 5.40 & 13.10 & 0.34 & 11.04 & 8.57 & 0.44 \\
    MS-BART & 7.45 & 9.66 & 0.44 & 10.99 & 8.31 & 0.51 \\
    DiffMS & 8.34 & 10.46 & 0.35 & 15.44 & 8.54 & 0.47 \\
    FRIGID-base & 19.80 & 7.88 & 0.54 & 24.41 & 6.20 & 0.61 \\
    FRIGID & 25.03 & \textbf{7.10} & 0.58 & 33.37 & 5.56 & 0.64 \\
    MS-GPT (published)$^{\ddagger}$ & \textbf{29.76} & 7.38 & \textbf{0.61} & 41.07 & \textbf{5.51} & 0.67 \\
    MS-GPT$^{\dagger}$ & 21.05 & 18.49 & 0.50 & 28.64 & 17.79 & 0.54 \\
    \midrule
    MS-GPT$^{\dagger}$ $\rightarrow$ GEMS & 26.28 & 20.61 & 0.52 & 42.22 & 19.54 & 0.60 \\
    FRIGID-base $\rightarrow$ GEMS & 28.02 & 11.38 & 0.56 & 43.34 & 10.23 & 0.64 \\
    GEMS (PubChem, Tani. $<0.7$) & 27.90 & 8.59 & 0.56 & 46.45 & 6.97 & 0.67 \\
    GEMS (one-edit starts excluded) & 25.53 & 8.69 & 0.55 & 42.22 & 7.06 & 0.66 \\
    \textbf{GEMS} & 29.64 & 7.73 & 0.59 & \textbf{50.81} & 6.08 & \textbf{0.71} \\
    \bottomrule
  \end{tabular}
\end{table}
\begin{table}[t]
  \caption{MassSpecGym known-formula baselines and GEMS variants. Non-MS-GPT published values come from MassSpecGym in the Wild Table~9 \citep{liu2026massspecgymwild}, with FRIGID-base from FRIGID Table~1 \citep{frigid2026}. $\dagger$: our best-effort same-split MS-GPT reproduction with corrected masking and retraining. $\ddagger$: published MS-GPT values are contextual comparisons with protocol qualifications in this section \citep{msgpt2026}. All GEMS variants use up to five starts, $P=B=16$, $D=3$, and $\lambda=16$. PubChem starts exclude the target connectivity. The one-edit row additionally excludes every target-specific one-edit neighbor connectivity group, without a Tanimoto filter. Generator and Tanimoto-filtered starts follow Table~\ref{tab:main_results}. Our evaluations retain all test queries. Best displayed values are bold, including ties.}
  \label{tab:published_massspecgym}
  \centering
  \scriptsize
  \setlength{\tabcolsep}{4pt}
  \begin{tabular}{lrrrrrr}
    \toprule
    Method & H@1 & MCES@1 & Tani.@1 & H@10 & MCES@10 & Tani.@10 \\
    \midrule
    MIST + MSNovelist & 0.00 & 39.84 & 0.06 & 0.00 & 18.83 & 0.15 \\
    Spec2Mol & 0.00 & 36.78 & 0.12 & 0.00 & 36.02 & 0.16 \\
    MIST + Neuraldecipher & 0.00 & 22.93 & 0.14 & 0.00 & 21.76 & 0.16 \\
    MADGEN & 1.31 & 27.47 & 0.20 & 1.54 & 16.84 & 0.26 \\
    MS-BART & 1.07 & 16.47 & 0.23 & 1.11 & 15.12 & 0.28 \\
    DiffMS & 2.30 & 13.96 & 0.28 & 4.25 & 11.68 & 0.39 \\
    FOAM (DiffMS + PubChem) & 1.50 & 12.21 & 0.35 & 10.28 & \textbf{6.06} & 0.53 \\
    MIST + MolForge (corrected) & 10.73 & 22.15 & 0.37 & 14.48 & 17.88 & 0.41 \\
    FRIGID-base & 16.09 & 15.02 & 0.41 & 18.19 & 11.95 & 0.46 \\
    FRIGID & 18.29 & 13.49 & 0.43 & 22.00 & 11.65 & 0.47 \\
    MS-GPT (published)$^{\ddagger}$ & 23.91 & 15.50 & 0.52 & 28.65 & 11.74 & 0.54 \\
    MS-GPT$^{\dagger}$ & 23.23 & 25.15 & 0.48 & 27.70 & 24.71 & 0.50 \\
    \midrule
    MS-GPT$^{\dagger}$ $\rightarrow$ GEMS & 25.56 & 26.66 & 0.48 & 34.07 & 25.91 & 0.52 \\
    GEMS (PubChem, Tani. $<0.7$) & 23.93 & 10.95 & 0.50 & 31.75 & 9.61 & 0.55 \\
    GEMS (one-edit starts excluded) & 22.44 & 11.02 & 0.49 & 29.57 & 9.65 & 0.55 \\
    \textbf{GEMS} & \textbf{30.22} & \textbf{9.68} & \textbf{0.55} & \textbf{41.33} & 8.26 & \textbf{0.61} \\
    \bottomrule
  \end{tabular}
\end{table}
\FloatBarrier

\FloatBarrier
\section{Learning the Navigation Policy}
\label{app:learning_controls}

\subsection{Pretraining and Supervision}
\label{sec:pretraining_control}

Table~\ref{tab:pretraining_control} compares random-weight and trained policies
under the same MIST-guided search and $\lambda=16$ ranking. The random-weight
policy uses the reported architecture without training and gives
\UntrainedHitOne\% Hit@1 and \UntrainedReach\% reach on NPLIB1 across three
weight initializations (mean $\pm$ sample standard deviation), well below every
trained NPLIB1 policy. Training improves performance under this fixed proposal and beam budget.

PubChem pretraining alone gives 28.89\% Hit@1 on NPLIB1 without downstream
fine-tuning, exceeding supervised training from scratch at 28.02\%, although
its Hit@10 and reach are lower. Fine-tuning adds 0.75 percentage points in
Hit@1, 3.99 in Hit@10, and 7.72 in reach. It therefore improves identification
as well as recovery. This contrast combines downstream training with
adaptation from exact target fingerprints to MIST predictions.

Pretraining followed by fine-tuning outperforms supervised training from
scratch in Hit@1, Hit@10, and reach on both benchmarks. The supervised models
use the same benchmark-specific navigation examples and validation-selected checkpoints.
From-scratch training uses six epochs on NPLIB1 and two on MassSpecGym,
with an initial learning rate of $10^{-4}$ and cosine decay to zero without
warmup on both benchmarks.

\begin{table}[htbp]
  \caption{Evaluation of random-weight and trained policies with $K=5$, $P=B=16$, $D=3$, MIST fingerprints for navigation and ranking, and $\lambda=16$. All 803 NPLIB1 and 17,556 MassSpecGym queries count. Pretrained only uses the PubChem checkpoint without downstream fine-tuning. Pretraining uses exact target fingerprints as inputs. REINFORCE uses MIST inputs and exact target fingerprints for its training reward. The random-weight row reports mean $\pm$ sample standard deviation across three weight initializations. Each trained setting uses one training random seed. Training schedules and selection details are given in the text.}
  \label{tab:pretraining_control}
  \centering
  \scriptsize
  \setlength{\tabcolsep}{4pt}
  \begin{tabular}{llrrr}
    \toprule
    Dataset & Policy & H@1 & H@10 & Reach \\
    \midrule
    NPLIB1 & Random weights & \ensuremath{4.61\pm2.90} & \ensuremath{7.97\pm4.98} & \ensuremath{8.30\pm5.29} \\
    NPLIB1 & RL from scratch (exact-FP reward) & 24.78 & 42.59 & 47.20 \\
    NPLIB1 & Supervised from scratch & 28.02 & 47.45 & 52.80 \\
    NPLIB1 & Pretrained only & 28.89 & 46.82 & 50.06 \\
    NPLIB1 & Pretrained + fine-tuned & 29.64 & 50.81 & 57.78 \\
    \midrule
    MassSpecGym & Supervised from scratch & 29.14 & 39.46 & 44.20 \\
    MassSpecGym & Pretrained + fine-tuned & 30.22 & 41.33 & 46.88 \\
    \bottomrule
  \end{tabular}
\end{table}
\FloatBarrier

\paragraph{Shortest-path supervision versus REINFORCE.}
\label{app:reinforce_control}
We compare training from scratch with shortest-path supervision against
REINFORCE using exact-target-fingerprint rewards on NPLIB1. The matched
recipe uses the same source--spectrum training examples, source-batch
schedule, global batch size of 64, and optimizer schedule. REINFORCE samples
32 three-edit trajectories per source. Rewards measure improvement in the
best Bernoulli agreement with the exact target fingerprint along a trajectory,
divided by 100, with a leave-one-out baseline across trajectories. Policy
inputs remain MIST predictions. The training schedule specifies 34,476
updates over six epochs, with four validation evaluations per epoch. Both
selected checkpoints are at step 27,293, with supervised checkpoint selection
based on $\lambda=16$ validation Hit@1.

Shortest-path supervision exceeds REINFORCE in Hit@1, Hit@10, and reach
(Table~\ref{tab:pretraining_control}). Matching the update and source-exposure
schedules does not match GPU-hours or establish
variability across training runs. The comparison does not isolate the effect
of retaining multiple optimal actions rather than one.

\subsection{Learned Search versus Enumerate \& Score}

Table~\ref{tab:direct_search} reports the accuracy and cost of each depth-three
configuration. At $P=B=4$, GEMS gives 27.65\% Hit@1 on NPLIB1 and 29.18\% on
MassSpecGym, with processing times of \NplibSmallPipelineSeconds{} and
\MsgSmallPipelineSeconds{} seconds. This setting costs less than width sixteen,
but recovers fewer targets. At the same beam width four, E\&S
reaches 54.17\% of NPLIB1 targets and 44.24\% of MassSpecGym targets, compared
with 49.07\% and 41.35\% for GEMS. GEMS has higher Hit@1 on both benchmarks
at this width and reduces processing time by factors of
\NplibMatchedWidthPipelineRatio{} and \MsgMatchedWidthPipelineRatio{},
respectively. E\&S has higher NPLIB1 Hit@10.

Increasing the learned beam to sixteen raises reach to 57.78\% and 46.88\%.
At this width, GEMS exceeds every tested E\&S width in Hit@1, Hit@10, and reach.
Compared with the widest tested E\&S beam ($B=4$), GEMS reduces reported
runtime by factors of \NplibDirectPipelineRatioRounded{} on NPLIB1 and
\MsgDirectPipelineRatioRounded{} on MassSpecGym.

\begin{table}[t]
  \caption{GEMS versus Enumerate \& Score (E\&S), with $K=5$, $D=3$ and $\lambda=16$. Each search task uses 32 physical CPU cores and 32 workers; GEMS also uses one H100. NPLIB1 GEMS at $P=B=16$ reports end-to-end wall time, including loading and output writing. Other runtime entries report accumulated processing time for preprocessing, retrieval, search, and final ranking (Appendix~\ref{app:pipeline_timing}). MassSpecGym search covers 16 tasks.}
  \label{tab:direct_search}
  \centering
  \scriptsize
  \setlength{\tabcolsep}{2.4pt}
  \begin{tabular}{@{}lrrrrr@{\hspace{10pt}}rrrrr@{}}
    \toprule
     & \multicolumn{5}{c}{\textbf{NPLIB1}} & \multicolumn{5}{c}{\textbf{MassSpecGym}} \\
    \cmidrule(lr){2-6}\cmidrule(lr){7-11}
    \shortstack{Search\\(Params.)} & H@1 & H@10 & Reach & \shortstack{Search\\(s)} & \shortstack{Runtime\\(s)} & H@1 & H@10 & Reach & \shortstack{Search\\(s)} & \shortstack{Runtime\\(s)} \\
    \midrule
    E\&S, $B=1$ & 25.40 & 43.84 & 44.21 & 770.01 & 802.88 & 28.03 & 37.98 & 38.57 & 25,250.45 & 25,351.17 \\
    E\&S, $B=2$ & 25.65 & 47.57 & 50.19 & 1,364.69 & 1,397.98 & 28.18 & 39.25 & 41.55 & 47,022.92 & 47,131.94 \\
    E\&S, $B=4$ & 26.03 & 49.07 & 54.17 & 2,491.07 & 2,524.59 & 28.16 & 39.05 & 44.24 & 87,247.68 & 87,372.32 \\
    GEMS, $P=B=4$ & 27.65 & 47.07 & 49.07 & 47.14 & 81.11 & 29.18 & 39.39 & 41.35 & 1,333.48 & 1,456.71 \\
    GEMS, $P=B=16$ & 29.64 & 50.81 & 57.78 & 129.28 & 200.12 & 30.22 & 41.33 & 46.88 & 3,869.90 & 4,091.00 \\
    \bottomrule
  \end{tabular}
\end{table}
\FloatBarrier

\FloatBarrier
\section{Where Search Starts}
\label{app:initialization}

\subsection{PubChem Starting Sets}
\label{app:initialization_sensitivity}
\label{app:longer_search}

Table~\ref{tab:initialization_depth} compares query-level identification and
recovery across four NPLIB1 starting sets at depths three, five, and ten.
The full-catalog and sampled-250 settings select up to five starts by MIST
agreement, using either the complete same-formula PubChem pool or a uniform
250-candidate draw. Random starts are the first up to 100 candidates of that
draw before fingerprint sorting. The fixed random-five condition uses their
first five starts. The same trained checkpoint and $P=B=16$ are used throughout.
MIST and exact-fingerprint conditions share the same starts, and each uses
its stated fingerprints for both navigation and ranking. Candidates and
consensus votes are restricted to the corresponding depth and starting set.

\begin{table}[htbp]
  \caption{NPLIB1 query-level Hit@1, Hit@10, and reach (\%) across initialization and depth, with $P=B=16$, $\lambda=16$, and the same trained checkpoint. All 803 queries count, including 17 without starts. MIST and Exact FP specify both navigation and ranking. Starts are fixed across guidance conditions. Random, up to 5 uses the first five starts of the random-100 searches. The ten-repeat random-five comparison is reported separately in the text.}
  \label{tab:initialization_depth}
  \centering
  \scriptsize
  \setlength{\tabcolsep}{3pt}
  \begin{tabular}{@{}lr@{\hspace{10pt}}rrr@{\hspace{12pt}}rrr@{}}
    \toprule
    \multirow{2}{*}{Initialization} & \multirow{2}{*}{$D$} & \multicolumn{3}{c}{\textbf{MIST}} & \multicolumn{3}{c}{\textbf{Exact FP (oracle)}} \\
    \cmidrule(lr){3-5}\cmidrule(lr){6-8}
     &  & H@1 & H@10 & Reach & H@1 & H@10 & Reach \\
    \midrule
    Full-catalog MIST, up to 5 & 3 & 29.64 & 50.81 & 57.78 & 68.99 & 72.73 & 72.85 \\
     & 5 & 29.76 & 50.56 & 58.78 & 73.23 & 77.58 & 77.96 \\
     & 10 & 29.64 & 50.81 & 59.90 & 79.45 & 84.43 & 84.93 \\
    \midrule
    Sampled-250 MIST, up to 5 & 3 & 26.15 & 43.46 & 50.44 & 61.15 & 64.38 & 64.51 \\
     & 5 & 27.90 & 46.20 & 53.80 & 68.99 & 73.23 & 73.60 \\
     & 10 & 27.90 & 48.19 & 56.54 & 77.58 & 82.57 & 83.06 \\
    \midrule
    Random, up to 5 (fixed) & 3 & 11.83 & 18.93 & 22.17 & 29.76 & 32.50 & 32.63 \\
     & 5 & 17.43 & 29.27 & 32.88 & 48.69 & 52.43 & 52.80 \\
     & 10 & 22.29 & 35.24 & 40.85 & 68.62 & 72.23 & 72.60 \\
    \midrule
    Random, up to 100 & 3 & 25.53 & 43.96 & 57.66 & 66.63 & 69.74 & 69.74 \\
     & 5 & 26.40 & 46.58 & 62.39 & 76.96 & 80.32 & 80.57 \\
     & 10 & 28.89 & 49.07 & 66.00 & 85.43 & 89.54 & 89.91 \\
    \bottomrule
  \end{tabular}
\end{table}
\FloatBarrier

\paragraph{Start quality.}
Under MIST guidance, full-catalog MIST selection gives the strongest D3
results among the tested initializations. Ranking within a 250-candidate
sample lowers Hit@1 from 29.64\% to 26.15\% and reach from 57.78\% to
50.44\%. Up to 100 random starts nearly match full-catalog reach at 57.66\%,
but identify fewer targets, with 25.53\% Hit@1.

\paragraph{Depth and identification.}
Deeper search partly compensates for weaker starts. At D10, the fixed
random-five setting achieves 22.29\% Hit@1 and 40.85\% reach, still below
the full-catalog ranked setting at D3. At D10, the random-100 setting reaches
more targets than the ranked setting (66.00\% versus 59.90\%), but identifies
fewer (28.89\% versus 29.64\% Hit@1).

\paragraph{Exact guidance.}
With exact fingerprints, the fixed random-five setting at D10 nearly matches
full-catalog ranked reach at D3 (72.60\% versus 72.85\%). Exact-guided search
from up to 100 random starts recovers the target for 89.91\% of queries by
D10. These starting sets therefore contain useful edit paths that MIST-guided
search does not fully exploit.

\paragraph{Repeated random-five selection.}
Across ten runs drawing up to five starts from the same fixed random-100
pool, D3 Hit@1 is $11.41\pm0.73\%$, Hit@10 is $20.40\pm0.74\%$, and reach
is $23.79\pm1.00\%$ (mean $\pm$ sample standard deviation). These runs use
different subsets from the fixed five-start prefix in
Table~\ref{tab:initialization_depth}.

\FloatBarrier
\subsection{Starts Filtered by Similarity to the Target}
\label{app:filtered_starts}

Target-excluded PubChem starts can still include close analogs of the target.
Motivated by the analysis of close analogs in pretraining data by
\citet{liu2026massspecgymwild}, we test sensitivity to similar structures
among inference-time starts. We exclude candidates whose Tanimoto similarity
to the target is at or above each cutoff, using radius-two, 2,048-bit Morgan
fingerprints after the molecular standardization described in
Appendix~\ref{app:structural_metrics}. In a separate condition, we exclude
all target-specific connectivity groups identified as one-edit neighbors
by exhaustive enumeration. Both conditions retain target-connectivity
exclusion. We then select up to five remaining candidates from the full
same-formula catalog by MIST agreement. The checkpoint, search settings,
and final ranking remain unchanged.

\begin{table}[b]
  \caption{Hit@1, Hit@10, and reach (\%) at $D=3$, with $P=B=16$, $\lambda=16$, and up to five MIST-ranked starts after exclusion. Tanimoto rows keep only starts below the cutoff. One-edit exclusion removes every starting connectivity one valid edit from the target. Target-connectivity exclusion applies throughout. MassSpecGym Tanimoto filtering was evaluated at cutoffs of 0.7 and above. All test queries count, including those without starts.}
  \label{tab:filtered_starts}
  \centering
  \scriptsize
  \begin{tabular}{lrrrrrr}
    \toprule
    & \multicolumn{3}{c}{NPLIB1} & \multicolumn{3}{c}{MassSpecGym} \\
    \cmidrule(lr){2-4}\cmidrule(lr){5-7}
    Starts removed & H@1 & H@10 & Reach & H@1 & H@10 & Reach \\
    \midrule
    Target only (reported) & 29.64 & 50.81 & 57.78 & 30.22 & 41.33 & 46.88 \\
    Tanimoto $\geq 0.9$ & 29.39 & 50.44 & 57.04 & 29.94 & 40.62 & 46.07 \\
    Tanimoto $\geq 0.8$ & 29.27 & 49.07 & 55.79 & 28.49 & 38.29 & 42.61 \\
    Tanimoto $\geq 0.7$ & 27.90 & 46.45 & 52.18 & 23.93 & 31.75 & 34.64 \\
    Tanimoto $\geq 0.6$ & 25.16 & 42.84 & 47.95 & -- & -- & -- \\
    Tanimoto $\geq 0.5$ & 21.92 & 36.36 & 40.10 & -- & -- & -- \\
    \midrule
    One edit from target & 25.53 & 42.22 & 46.58 & 22.44 & 29.57 & 32.03 \\
    \bottomrule
  \end{tabular}
\end{table}

\paragraph{Results.}
Accuracy falls as stricter similarity cutoffs are applied
(Table~\ref{tab:filtered_starts}). At a cutoff of 0.7, the threshold used
for close analogs in \emph{MassSpecGym in the Wild}
\citep{liu2026massspecgymwild}, GEMS gives 27.90\%/46.45\% Hit@1/Hit@10 on
NPLIB1 and 23.93\%/31.75\% on MassSpecGym, above FRIGID's
25.03\%/33.37\% and 18.29\%/22.00\%
(Table~\ref{tab:main_results}). MassSpecGym is more sensitive to the filter,
losing about 6.3 Hit@1 points at 0.7 compared with 1.7 on NPLIB1. On NPLIB1,
Hit@1 falls below FRIGID's only at a cutoff of 0.5 (21.92\%), while Hit@10
remains higher (36.36\% versus 33.37\%). At this cutoff, additional edit
rounds raise NPLIB1 Hit@1 from 9.34\% at $D=1$ to 19.05\% at $D=2$ and
21.92\% at $D=3$.

With up to five MIST-ranked starts and $P=B=16$, the D1/D2/D3 Hit@1
values are 27.02\%/29.39\%/29.64\% on NPLIB1 and
30.09\%/30.43\%/30.22\% on MassSpecGym without Tanimoto filtering.
At cutoff 0.7, they are 22.79\%/27.27\%/27.90\% and
18.75\%/22.76\%/23.93\%, respectively. The D1--D2 gains are therefore
4.48 and 4.02 percentage points after filtering, compared with 2.37 and
0.34 without it; the D2--D3 gains are 0.62 and 1.17, compared with
0.25 and $-0.21$. Gains are computed from hit counts before rounding.

With every one-edit neighbor removed, no target is recovered at $D=1$.
Every target recovered by $D=3$ is therefore reached through two or three
edits. GEMS gives 25.53\%/42.22\% Hit@1/Hit@10 on NPLIB1 and
22.44\%/29.57\% on MassSpecGym, still above FRIGID on both benchmarks.
Increasing depth from two to three raises
Hit@1 from 24.16\% to 25.53\% on NPLIB1 and from 20.31\% to 22.44\% on
MassSpecGym.
With target exclusion alone, the corresponding Hit@1 values change from
29.39\% to 29.64\% on NPLIB1 and from 30.43\% to 30.22\% on MassSpecGym.
These depth comparisons use up to five starts and $P=B=16$.

Additional starts give larger identification gains under stricter similarity
filtering. Increasing the maximum number of starts from five to twenty raises
NPLIB1 Hit@1 by 0.75 points without filtering and by 2.12 points at a cutoff
of 0.5, and Hit@10 by 1.99 and 5.35 points, respectively. Reach rises by
about eight to nine points at every cutoff.

\FloatBarrier
\subsection{Generated Starting Pools and Refinement}
\label{app:generator_refinement}
\suppressfloats[t]

For both FRIGID and MS-GPT, we select up to five executable same-formula
candidates by MIST Bernoulli score, retaining the target when present.
The original-ranking rows in Table~\ref{tab:generator_refinement} preserve
each generator's output order. Selected (D0) evaluates the chosen
starts, and GEMS (D3) adds states retained during three edit rounds from those
same starts. Both use MIST fingerprints and $\lambda=16$ ranking, with
$P=B=16$ for search. All 803 NPLIB1 and 17,556 MassSpecGym queries remain
included, counting queries without executable starts as misses.

\paragraph{FRIGID candidate pools.}
The NPLIB1 FRIGID-base pool is the initial round of the same run that produces
the final FRIGID pool. Its 65,151 candidates precede all ICEBERG
refinement. FRIGID selects up to twelve parents by its own ranking in each
of 24 refinement rounds, producing eight masked variants per parent and
adding 32 fresh samples. Missing refinement slots are filled with fresh
samples. GEMS and FRIGID therefore share the initial generated pool, while
parent selection, ranking, and computational budgets differ.
For further refinement, GEMS selects from the top 128 saved candidates per
query after the 24 FRIGID refinement rounds. The original FRIGID row reports
identification accuracy only because the full cumulative pool was not saved.
There are \NplibFrigidZeroSeedMisses{} NPLIB1 queries without executable
FRIGID-base starts and \NplibIcebergZeroSeedMisses{} without executable
refined FRIGID starts. MassSpecGym includes 474 queries for which FRIGID-base outputs are
unavailable.

\paragraph{MS-GPT candidate pools.}
We use outputs from our best-effort corrected and retrained MS-GPT
reproduction (Appendix~\ref{app:structural_metrics}), applying the same
MIST-based selection and GEMS refinement procedure. There are 128 NPLIB1
and 3,633 MassSpecGym queries without executable selected starts.

\paragraph{Proximity of generated starts.}
Table~\ref{tab:frigid_seed_proximity} compares the three NPLIB1 starting
pools. D0 checks whether a selected start already matches the target, and
D1 exhaustively enumerates every valid one-edit product. D2/D3 combine this
coverage with saved MIST- and oracle-guided paths, giving lower bounds on
the fraction of queries with a start within that many edits.
Among queries without a correct selected start, three-edit coverage is at
least 292/611 (47.79\%) for FRIGID-base, 275/561 (49.02\%) for refined FRIGID,
and 240/582 (41.24\%) for MS-GPT. Queries without starts remain included.
These conditional sets differ between generators. The counts establish
proximity of the selected starts, independently of final ranking.

\begin{table}[t]
  \caption{NPLIB1 queries with at least one selected start within D edits of the target. Each generator pool supplies up to five MIST-selected starts. All 803 queries count, including those without starts. D0/D1 are exact. D2/D3 are lower bounds combining exhaustive one-edit coverage with saved MIST- and oracle-guided paths. MS-GPT (ours) is our best-effort reproduction. These counts measure proximity, independently of final ranking.}
  \label{tab:frigid_seed_proximity}
  \centering
  \scriptsize
  \setlength{\tabcolsep}{4pt}
  \begin{tabular}{lrrrrrr}
    \toprule
     & \multicolumn{2}{c}{FRIGID-base} & \multicolumn{2}{c}{FRIGID after refinement} & \multicolumn{2}{c}{MS-GPT (ours)} \\
    \cmidrule(lr){2-3}\cmidrule(lr){4-5}\cmidrule(lr){6-7}
    Maximum edits & Queries & \% & Queries & \% & Queries & \% \\
    \midrule
    D0 & 192 & 23.91 & 242 & 30.14 & 221 & 27.52 \\
    D1 & 319 & 39.73 & 377 & 46.95 & 344 & 42.84 \\
    D2 & $\geq$420 & $\geq$52.30 & $\geq$453 & $\geq$56.41 & $\geq$410 & $\geq$51.06 \\
    D3 & $\geq$484 & $\geq$60.27 & $\geq$517 & $\geq$64.38 & $\geq$461 & $\geq$57.41 \\
    \bottomrule
  \end{tabular}
\end{table}

\paragraph{Refinement results.}
Table~\ref{tab:generator_refinement} compares the original generator rankings,
selected starts, and GEMS refinement. Three edit rounds improve Hit@1,
Hit@10, and reach over the selected-start baseline for every evaluated
generator pool. On NPLIB1, GEMS exceeds our FRIGID run's original Hit@1 of
24.03\%, reaching 28.02\% from its initial pool and 28.27\% from its refined
pool. GEMS also improves on the original ranking of our best-effort MS-GPT
reproduction, raising Hit@1 from 21.05\% to 26.28\% on NPLIB1 and from
23.23\% to 25.56\% on MassSpecGym. These results use MIST predictions for
both navigation and ranking.

\begin{table}[htbp]
  \caption{Generator outputs and GEMS refinement. FRIGID-base and FRIGID are the initial and final pools of the same NPLIB1 run. MS-GPT (ours) is our best-effort reproduction with corrected masking and retraining. Original ranking uses the generator output order. Selected (D0) and GEMS (D3) use the same up-to-five MIST-ranked starts and $\lambda=16$. All 803 NPLIB1 and 17,556 MassSpecGym queries count. Oracle results are separate in Table~\ref{tab:msgpt_refinement_oracle}.}
  \label{tab:generator_refinement}
  \centering
  \scriptsize
  \setlength{\tabcolsep}{4pt}
  \begin{tabular}{lllrrr}
    \toprule
    Dataset & Initializer & Stage & H@1 & H@10 & Reach \\
    \midrule
    NPLIB1 & FRIGID-base & Selected (D0) & 20.42 & 23.91 & 23.91 \\
    NPLIB1 & FRIGID-base & GEMS (D3) & 28.02 & 43.34 & 46.20 \\
    NPLIB1 & FRIGID & Original ranking & 24.03 & 33.13 & -- \\
    NPLIB1 & FRIGID & Selected (D0) & 22.54 & 30.14 & 30.14 \\
    NPLIB1 & FRIGID & GEMS (D3) & 28.27 & 46.82 & 50.31 \\
    NPLIB1 & MS-GPT (ours) & Original ranking & 21.05 & 28.64 & -- \\
    NPLIB1 & MS-GPT (ours) & Selected (D0) & 22.67 & 27.52 & 27.52 \\
    NPLIB1 & MS-GPT (ours) & GEMS (D3) & 26.28 & 42.22 & 45.08 \\
    \midrule
    MassSpecGym & FRIGID-base & Selected (D0) & 11.60 & 12.16 & 12.16 \\
    MassSpecGym & FRIGID-base & GEMS (D3) & 18.12 & 23.43 & 24.72 \\
    MassSpecGym & MS-GPT (ours) & Original ranking & 23.23 & 27.70 & -- \\
    MassSpecGym & MS-GPT (ours) & Selected (D0) & 24.00 & 26.68 & 26.68 \\
    MassSpecGym & MS-GPT (ours) & GEMS (D3) & 25.56 & 34.07 & 36.64 \\
    \bottomrule
  \end{tabular}
\end{table}
\FloatBarrier

\FloatBarrier
\subsection{Oracle Guidance from Generated Starts}
Table~\ref{tab:msgpt_refinement_oracle} replaces navigation and ranking evidence
with exact target fingerprints for the FRIGID and MS-GPT starts. Seed selection still
uses MIST, so the starting pools are identical to the corresponding
refinement runs. These are oracle diagnostics. Oracle D3 results are prefixes
of completed ten-round searches, as are the corresponding MIST results for
MS-GPT. The FRIGID MIST comparisons use standalone three-round searches.

With the starting pools fixed, exact fingerprints improve recovery and
identification from both FRIGID pools on NPLIB1 and from MS-GPT on both
benchmarks. On NPLIB1, Hit@1
rises from 28.02\% to 57.91\% for FRIGID-base and from 28.27\% to 61.52\%
for refined FRIGID. Navigation and ranking change together in these
comparisons. FRIGID-base oracle reach is 60.15\%, while the 60.27\% proximity
bound in Table~\ref{tab:frigid_seed_proximity} also includes exhaustive
one-edit coverage and paths witnessed by other searches.

Generated starting pools often contain candidates connected to the target by
short edit paths. However, proximity alone does not ensure successful refinement:
MIST-guided search misses some reachable targets, and final ranking does not
consistently place recovered targets first. Exact-fingerprint interventions
improve both recovery and identification from the same starting pools.

\begin{table}[t]
  \caption{Oracle diagnostics for generated starts. Exact target fingerprints guide navigation and ranking, using the same MIST-selected starts as Table~\ref{tab:generator_refinement}. Search uses up to $K=5$ starts, $P=B=16$, $D=3$, and $\lambda=16$. MS-GPT (ours) is our best-effort reproduction. All 803 NPLIB1 and 17,556 MassSpecGym queries count, including missing starts.}
  \label{tab:msgpt_refinement_oracle}
  \centering
  \scriptsize
  \setlength{\tabcolsep}{4pt}
  \begin{tabular}{lllrrr}
    \toprule
    Dataset & Initializer & Stage & H@1 & H@10 & Reach \\
    \midrule
    NPLIB1 & FRIGID-base & GEMS (D3) & 57.91 & 59.90 & 60.15 \\
    NPLIB1 & FRIGID & GEMS (D3) & 61.52 & 64.38 & 64.38 \\
    NPLIB1 & MS-GPT (ours) & Selected (D0) & 27.27 & 27.52 & 27.52 \\
    NPLIB1 & MS-GPT (ours) & GEMS (D3) & 55.17 & 57.41 & 57.41 \\
    \midrule
    MassSpecGym & MS-GPT (ours) & Selected (D0) & 26.13 & 26.68 & 26.68 \\
    MassSpecGym & MS-GPT (ours) & GEMS (D3) & 41.05 & 42.87 & 42.92 \\
    \bottomrule
  \end{tabular}
\end{table}
\FloatBarrier

\section{Search Width and Depth}
\label{app:depth_analysis}

\subsection{Width versus Depth}
\label{app:width_depth}

\begin{figure}[htbp]
  \centering
  \includegraphics[width=\linewidth]{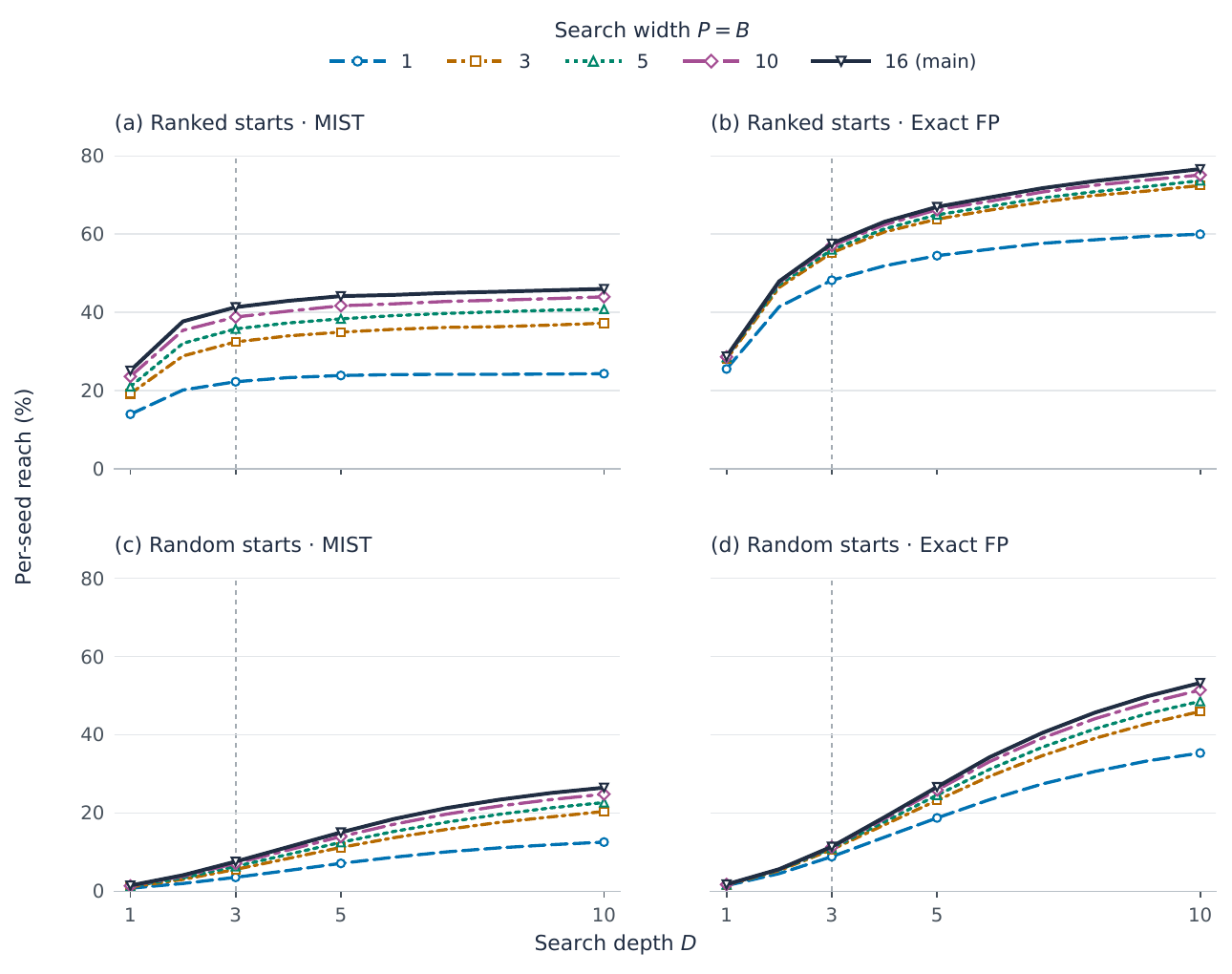}
\caption{\textbf{Width and depth affect recovery differently.}
Per-seed reach on NPLIB1 from up to five MIST-ranked starts or up to 100 random
starts per query. Each curve uses $P=B$ and the same trained checkpoint.
The vertical line marks the main depth $D=3$. Exact FP uses fingerprints
computed from the known targets and is an oracle diagnostic. Denominators
are 3,888 ranked and 68,811 random seed--target pairs. Queries without starts
contribute no pairs. Table~\ref{tab:width_depth} gives the numerical values.}
  \label{fig:width_depth}
\end{figure}

The NPLIB1 grid varies $P=B$ and depth using the reported model and fixed
ordered starting sets: up to five MIST-ranked starts or up to 100 random
starts per query. Depths are prefixes of ten-round searches.
Table~\ref{tab:width_depth_query} reports query-level identification, while
Figure~\ref{fig:width_depth} and Table~\ref{tab:width_depth} report per-seed
reach, the fraction of individual seed--target searches that recover the
target. Per-seed reach separates individual search success from the number
of opportunities per query, although queries with more starts receive more
weight and the starting distributions differ.

\paragraph{Width versus depth under MIST.}
From the reference setting $P=B=D=3$, we compare widening to $P=B=10$ at
$D=3$ with deepening to $D=10$ at $P=B=3$. Both expanded settings retain at
most $BD=30$ search states plus the initial structure per seed before final
deduplication. This matches the maximum retained-candidate budget, while
actual unique candidate counts and computation can differ.
From ranked starts, widening raises Hit@1 from 28.02\% to 29.89\%, compared
with 29.02\% for deepening. Per-seed reach similarly rises from 32.43\% to
38.79\% with widening and 37.22\% with deepening. From random starts, the
ordering reverses: widening raises Hit@1 from 24.91\% to 25.90\%, while
deepening raises it to 29.39\%. Per-seed reach rises from 5.58\% to 7.10\%
and 20.38\%, respectively. Hit@10 follows the same ordering for both starting
sets. Ranked starts therefore benefit more from exploring alternatives
within a short search, whereas random starts benefit more from extending
trajectories in this comparison.

\paragraph{Exact fingerprint evidence changes the tradeoff.}
With exact fingerprints, deepening is more beneficial for both starting
sets. From ranked starts, reference Hit@1 is 68.99\%, compared with 68.87\%
for widening and 78.46\% for deepening. From random starts, the corresponding
values are 66.25\%, 66.50\%, and 84.18\%. Per-seed reach and Hit@10 support
the same ordering. Deeper search can thus yield substantially greater gains
with more informative fingerprint evidence. Exact fingerprints replace
both navigation and ranking inputs here, so this comparison does not
isolate their individual contributions.

\paragraph{Diminishing returns in identification.}
From ranked starts under MIST, widths ten and sixteen differ by at most
0.50 percentage points in Hit@1 at the reported depths $D\in\{1,3,5,10\}$.
Recovery still improves: at $D=3$, per-seed reach rises from 38.79\% to
41.31\% between these widths. At width sixteen, extending $D=3$ to $D=10$
further raises per-seed reach to 45.99\%, while Hit@1 and Hit@10 remain at
29.64\% and 50.81\% at the two endpoints. Additional search therefore
continues to improve recovery without necessarily improving identification.

\begin{table}[htbp]
  \caption{Query-level Hit@1 / Hit@10 (\%) on NPLIB1 for the width--depth settings in Table~\ref{tab:width_depth}. All 803 queries count, including 17 without starts. Each run uses $P=B$ and $\lambda=16$ consensus with initial-seed support. Fingerprints specify both navigation and ranking. Exact FP uses the known target structure and is an oracle diagnostic. Candidates and consensus votes are restricted to depth $D$.}
  \label{tab:width_depth_query}
  \centering
  \scriptsize
  \setlength{\tabcolsep}{3pt}
  \begin{tabular}{llrrrrr}
    \toprule
    Initialization & Fingerprints & $P=B$ & $D=1$ & $D=3$ & $D=5$ & $D=10$ \\
    \midrule
    MIST-ranked, up to 5 & MIST & 1 & 23.54 / 32.38 & 25.16 / 36.74 & 25.78 / 37.24 & 26.03 / 37.48 \\
     &  & 3 & 26.03 / 40.35 & 28.02 / 45.58 & 28.39 / 45.95 & 29.02 / 47.07 \\
     &  & 5 & 26.15 / 43.09 & 28.52 / 47.95 & 28.89 / 48.32 & 28.89 / 48.19 \\
     &  & 10 & 26.65 / 45.70 & 29.89 / 49.56 & 29.76 / 50.06 & 30.14 / 50.56 \\
     &  & 16 & 27.02 / 46.82 & 29.64 / 50.81 & 29.76 / 50.56 & 29.64 / 50.81 \\
    \midrule
    MIST-ranked, up to 5 & Exact FP & 1 & 54.55 / 55.29 & 66.00 / 68.12 & 68.99 / 71.61 & 74.10 / 76.71 \\
     &  & 3 & 56.91 / 57.66 & 68.99 / 71.48 & 72.35 / 75.47 & 78.46 / 81.82 \\
     &  & 5 & 57.16 / 58.41 & 68.74 / 71.36 & 73.10 / 75.97 & 79.45 / 82.57 \\
     &  & 10 & 57.53 / 58.78 & 68.87 / 72.23 & 73.10 / 77.21 & 78.83 / 83.56 \\
     &  & 16 & 57.66 / 58.90 & 68.99 / 72.73 & 73.23 / 77.58 & 79.45 / 84.43 \\
    \midrule
    Random, up to 100 & MIST & 1 & 15.57 / 23.91 & 23.04 / 35.99 & 25.65 / 39.10 & 27.27 / 42.47 \\
     &  & 3 & 16.56 / 29.51 & 24.91 / 41.34 & 26.90 / 43.84 & 29.39 / 46.45 \\
     &  & 5 & 16.94 / 30.64 & 25.40 / 42.59 & 26.77 / 44.33 & 28.64 / 46.95 \\
     &  & 10 & 17.93 / 32.38 & 25.90 / 42.84 & 26.65 / 46.08 & 29.14 / 48.19 \\
     &  & 16 & 18.31 / 32.88 & 25.53 / 43.96 & 26.40 / 46.58 & 28.89 / 49.07 \\
    \midrule
    Random, up to 100 & Exact FP & 1 & 43.21 / 44.33 & 63.64 / 64.76 & 71.73 / 73.35 & 80.57 / 82.44 \\
     &  & 3 & 45.58 / 46.70 & 66.25 / 68.37 & 75.22 / 78.08 & 84.18 / 87.30 \\
     &  & 5 & 45.58 / 47.20 & 66.25 / 68.74 & 75.84 / 79.08 & 84.81 / 88.29 \\
     &  & 10 & 45.83 / 47.32 & 66.50 / 68.99 & 76.46 / 80.07 & 85.06 / 89.17 \\
     &  & 16 & 45.95 / 47.32 & 66.63 / 69.74 & 76.96 / 80.32 & 85.43 / 89.54 \\
    \bottomrule
  \end{tabular}
\end{table}
\begin{table}[htbp]
  \caption{Per-seed reach (\%) on NPLIB1 for the searches in Figure~\ref{fig:width_depth}. A seed--target pair succeeds when its own retained lineage contains the target by depth $D$. Denominators are 3,888 pairs for MIST-ranked starts and 68,811 pairs for random starts. Queries without starts contribute no pairs. All runs use $P=B$ and the same trained checkpoint. Exact FP is an oracle diagnostic.}
  \label{tab:width_depth}
  \centering
  \scriptsize
  \setlength{\tabcolsep}{4pt}
  \begin{tabular}{llrrrrr}
    \toprule
    Starts & Guidance & $P=B$ & $D=1$ & $D=3$ & $D=5$ & $D=10$ \\
    \midrule
    MIST-ranked, up to 5 & MIST & 1 & 13.94 & 22.25 & 23.84 & 24.31 \\
     &  & 3 & 19.19 & 32.43 & 34.93 & 37.22 \\
     &  & 5 & 21.01 & 35.75 & 38.32 & 40.82 \\
     &  & 10 & 23.61 & 38.79 & 41.67 & 43.96 \\
     &  & 16 & 25.10 & 41.31 & 44.14 & 45.99 \\
    \midrule
    MIST-ranked, up to 5 & Exact FP & 1 & 25.51 & 48.20 & 54.48 & 59.95 \\
     &  & 3 & 28.01 & 55.25 & 63.81 & 72.43 \\
     &  & 5 & 28.47 & 55.99 & 64.92 & 73.66 \\
     &  & 10 & 28.60 & 57.02 & 66.23 & 75.10 \\
     &  & 16 & 28.73 & 57.56 & 66.95 & 76.62 \\
    \midrule
    Random, up to 100 & MIST & 1 & 0.76 & 3.55 & 7.12 & 12.56 \\
     &  & 3 & 1.08 & 5.58 & 11.18 & 20.38 \\
     &  & 5 & 1.21 & 6.33 & 12.49 & 22.67 \\
     &  & 10 & 1.36 & 7.10 & 13.97 & 24.81 \\
     &  & 16 & 1.46 & 7.57 & 15.04 & 26.47 \\
    \midrule
    Random, up to 100 & Exact FP & 1 & 1.44 & 8.84 & 18.73 & 35.33 \\
     &  & 3 & 1.62 & 10.64 & 23.23 & 45.96 \\
     &  & 5 & 1.67 & 10.98 & 24.55 & 48.55 \\
     &  & 10 & 1.72 & 11.26 & 25.89 & 51.42 \\
     &  & 16 & 1.74 & 11.39 & 26.65 & 53.25 \\
    \bottomrule
  \end{tabular}
\end{table}
\FloatBarrier

\subsection{Recovery by Depth on Both Benchmarks}
\label{app:seed_proximity_evidence}

Table~\ref{tab:witnessed_request_paths} reports query-level recovery by depth
on both benchmarks, using the reported trained policies, up to five
MIST-ranked PubChem starts, and $P=B=16$. The retained paths show that many
target-excluded retrieved isomers are close to an accepted target. With
MIST guidance, at least one start reaches the target within a single edit
for 51.56\% of NPLIB1 queries and 43.30\% of MassSpecGym queries. Recovery
rises to \NplibReach\% and \MsgReach\% by depth three. These measurements
depend on the candidate pool, independently of final ranking.

On both benchmarks, additional depth yields substantially larger recovery
gains with exact fingerprint evidence. Extending search from D3 to D10
adds 2.12 percentage points of reach on NPLIB1 and 0.94 on MassSpecGym under
MIST guidance. With exact fingerprints, the same extension adds 12.08 and
8.96 points.

Additional recovery does not consistently improve top-ranked identification.
We compare depth-three and depth-five prefixes of the same searches, using
the candidates and lineage votes retained through each depth. Under MIST
guidance, extending search from D3 to D5 raises NPLIB1 reach from 57.78\%
to 58.78\%, while Hit@1 changes from 29.64\% to 29.76\%. On MassSpecGym,
reach rises from 46.88\% to 47.30\%, while Hit@1 decreases from 30.22\%
to 29.88\%.

\begin{table}[t]
  \caption{Queries with a witnessed path of at most D edits from at least one MIST-ranked starting isomer. All queries remain in the denominator, including queries without starts. Percentages are recovery, independent of final ranking. These provide lower bounds on the fraction of queries having a seed within D edits. Failures do not establish a larger shortest distance. Exact FP is an oracle diagnostic.}
  \label{tab:witnessed_request_paths}
  \centering
  \scriptsize
  \setlength{\tabcolsep}{4pt}
  \begin{tabular}{llrrrrrr}
    \toprule
    Dataset & Guidance & Queries & D1 & D2 & D3 & D5 & D10 \\
    \midrule
    NPLIB1 & MIST & 803 & 51.56 & 56.66 & 57.78 & 58.78 & 59.90 \\
    NPLIB1 & Exact FP & 803 & 59.03 & 69.49 & 72.85 & 77.96 & 84.93 \\
    MassSpecGym & MIST & 17556 & 43.30 & 45.94 & 46.88 & 47.30 & 47.82 \\
    MassSpecGym & Exact FP & 17556 & 48.18 & 53.86 & 56.07 & 58.40 & 65.03 \\
    \bottomrule
  \end{tabular}
\end{table}
\FloatBarrier

\subsection{Witnessed Paths and Certified Distances}
\label{app:certified_navigation}

A recovered target establishes a valid edit path, but the first recovery
depth need not be the shortest possible distance. Certifying that distance
requires both a verified path and exhaustive exclusion of shorter paths.
Figure~\ref{fig:overview} illustrates a certified three-edit example.
Table~\ref{tab:witnessed_request_paths} reports witnessed paths across the
benchmarks, where this certification procedure has not been applied to every
starting seed. Failure to recover a target with a finite beam leaves its
distance unresolved.

We measure $d_{\mathrm{eval}}(s,t)$ from the exact initial seed state, accepting
any endpoint whose InChIKey14 matches the reference target. For a
 target-excluded seed, exhaustive forward search through two edits tests every
reachable state for a match. A first match at depth one or two certifies that
distance. If neither depth contains a match, the distance is at least three.

To search for a three-step path, we also expand the reference target by one
edit. The InChIKey14 values in that neighborhood select promising
seed-reachable states at depth two. We then expand those states forward from
their actual bond-and-hydrogen configurations. A path is accepted only when
its endpoint matches the target InChIKey14 and all three edits replay
successfully. Combined with the two-step lower bound, this certifies distance
three.

The target-side filter can miss paths ending at other accepted target
structures. When no verified path is found, the lower bound remains three
and the exact distance is unresolved.

\FloatBarrier

\section{Fingerprint Evidence in Navigation and Ranking}
\label{app:fingerprint_ablation}

\subsection{Fingerprint Interventions with PubChem Starts}

Table~\ref{tab:fingerprint_oracle} reports fingerprint interventions with and
without consensus. Its $\lambda=16$ rows give the numerical results for
Figure~\ref{fig:fingerprint_oracle}.

\begin{table}[t]
  \caption{Fingerprint replacement during navigation and ranking at $K=5$, $P=B=16$, and $D=3$, with pure Bernoulli ranking ($\lambda=0$) or consensus ($\lambda=16$). All rows use the same MIST-selected starts. Changing only the ranking fingerprint preserves the candidate pool and reach. Metrics are percentages. MassSpecGym includes both adducts. Rows with $\lambda=16$ give the values in Figure~\ref{fig:fingerprint_oracle}.}
  \label{tab:fingerprint_oracle}
  \label{tab:gems_results}
  \label{tab:oracle_full}
  \centering
  \scriptsize
  \setlength{\tabcolsep}{4pt}
  \begin{tabular}{lllrrrrrr}
    \toprule
     &  &  & \multicolumn{3}{c}{\textbf{NPLIB1}} & \multicolumn{3}{c}{\textbf{MassSpecGym}} \\
    \cmidrule(lr){4-6}\cmidrule(lr){7-9}
    Navigation & Ranking & $\lambda$ & H@1 & H@10 & Reach & H@1 & H@10 & Reach \\
    \midrule
    MIST & MIST & 0 & 28.02 & 50.56 & 57.78 & 29.47 & 40.94 & 46.88 \\
    MIST & MIST & 16 & 29.64 & 50.81 & 57.78 & 30.22 & 41.33 & 46.88 \\
    MIST & Exact FP & 0 & 55.54 & 57.66 & 57.78 & 44.76 & 46.74 & 46.88 \\
    MIST & Exact FP & 16 & 56.66 & 57.53 & 57.78 & 45.12 & 46.79 & 46.88 \\
    \midrule
    Exact FP & MIST & 0 & 29.39 & 58.78 & 72.85 & 31.19 & 47.10 & 56.07 \\
    Exact FP & MIST & 16 & 33.25 & 61.89 & 72.85 & 33.82 & 49.74 & 56.07 \\
    Exact FP & Exact FP & 0 & 67.62 & 72.60 & 72.85 & 51.04 & 55.87 & 56.07 \\
    Exact FP & Exact FP & 16 & 68.99 & 72.73 & 72.85 & 53.09 & 55.94 & 56.07 \\
    \bottomrule
  \end{tabular}
\end{table}
\FloatBarrier

\paragraph{Navigation.}
With $\lambda=16$ and MIST ranking, replacing the navigation fingerprint with
the exact target fingerprint raises reach from 57.78\% to 72.85\% on NPLIB1
and from 46.88\% to 56.07\% on MassSpecGym. Hit@1 increases from 29.64\% to
33.25\% and from 30.22\% to 33.82\%, respectively. Recovery therefore
improves more than top-ranked identification.

\paragraph{Ranking.}
Many recovered targets are not placed first by MIST ranking. With the
MIST-guided candidate pools fixed, replacing only the ranking fingerprint
raises Hit@1 from 29.64\% to 56.66\% on NPLIB1 and from 30.22\% to 45.12\%
on MassSpecGym. These values approach the corresponding reach of 57.78\%
and 46.88\%. Candidate pools and consensus counts remain unchanged, so the
improvement comes from selecting among already recovered candidates rather
than additional search.

Replacing both navigation and ranking fingerprints raises Hit@1 further,
to 68.99\% on NPLIB1 and 53.09\% on MassSpecGym. Together, these oracle
interventions show scope to improve both target recovery and selection
from the recovered pool, with the starting sets and trained checkpoints
held fixed.

\subsection{Effect of Consensus}
\label{sec:ranking_results}

We select $\lambda=\ConsensusWeight{}$ by MassSpecGym validation Hit@1 across
both adducts and use it for both benchmarks.

Consensus reorders a fixed candidate pool without changing reach. It raises
Hit@1 from \NplibFPHitOne\% to \NplibHitOne\% on NPLIB1 and from
\MsgFPHitOne\% to \MsgHitOne\% on MassSpecGym
(Table~\ref{tab:gems_results}). The improvement in NPLIB1 Hit@10 is smaller,
from 50.56\% to 50.81\%.

Among the tested interventions, consensus gives its largest Hit@1 gains with
exact navigation and MIST ranking: 3.86 percentage points on NPLIB1 and 2.63
on MassSpecGym.

Removing consensus leaves the main conclusion of the fingerprint experiment
unchanged (Table~\ref{tab:oracle_full}). With the MIST-guided pools fixed,
exact-fingerprint ranking raises Hit@1 from 28.02\% to 55.54\% on NPLIB1 and
from 29.47\% to 44.76\% on MassSpecGym. Using exact fingerprints for both
navigation and ranking raises these values to 67.62\% and 51.04\%.
These comparisons use the same seeds, checkpoint, and candidate pools as the
$\lambda=16$ conditions. Only the consensus term is removed.

\subsection{Reranking with Predicted Spectra}
\label{app:iceberg_reranking}

We test whether predicted spectra provide complementary ranking evidence by
reranking the fixed top-ten lists from the main GEMS configuration. ICEBERG
\citep{goldman2024iceberg} predicts a spectrum for each candidate using the
benchmark-specific checkpoints released with FRIGID. We add a spectral score
$S(m)$ to the GEMS score, giving $R(m;q)+\beta S(m)$, while keeping
$\lambda=16$. The metric, energy aggregation, and $\beta$ are selected by
validation Hit@1, with $\beta\in\{0,4,8,16,32,64,128\}$.

NPLIB1 selects $\beta=128$ and experimental intensity coverage, the fraction
of observed intensity matched by predicted peaks. MassSpecGym selects
$\beta=32$ and cosine similarity of square-root intensities. Both use
one-to-one peak matching within 20\,ppm and exclude precursor and
above-precursor peaks. For NPLIB1, scores use matched recorded collision
energies when available and otherwise average over 10--90\,eV in 10\,eV
increments. MassSpecGym uses the nearest grid energy to the recorded value,
or 40\,eV when energy is unavailable.

\begin{table}[htbp]
  \caption{ICEBERG reranking of the fixed GEMS top-ten lists. Hit@1 is reported in percent. Gains and their paired 95\% bootstrap intervals, grouped by target connectivity, are in percentage points.}
  \label{tab:iceberg_reranking}
  \centering
  \small
  \begin{tabular}{lrr}
    \toprule
    Ranking & NPLIB1 & MassSpecGym \\
    \midrule
    GEMS & 29.64 & 30.22 \\
    GEMS + ICEBERG & 31.88 & 31.60 \\
    \midrule
    Gain [95\% interval] & +2.24 [0.62, 3.96] & +1.38 [0.40, 2.80] \\
    \bottomrule
  \end{tabular}
\end{table}

The validation-selected spectral scores improve Hit@1 on both benchmarks
(Table~\ref{tab:iceberg_reranking}), showing that forward spectral predictions
complement fingerprint-based ranking. Substantial identification error
remains within the fixed candidate lists. All test queries remain in the
denominators. Hit@10 and full-pool reach are unchanged because reranking
preserves candidate membership and cannot rescue targets outside the top ten.
Appendix~\ref{app:iceberg_timing} reports the additional prediction cost.

\FloatBarrier
\section{Runtime Measurements}
\label{app:runtime}

At $K=5$, $P=B=16$, and $D=3$, GEMS takes \NplibPipelineSeconds{} seconds
end to end on NPLIB1, the median of three runs over all 803 test spectra,
or \NplibPipelinePerSpectrum{} seconds per spectrum. This wall time includes
process startup, model and data loading, spectrum preprocessing, peak
annotation, MIST prediction, fresh fingerprint database construction,
retrieval, search, final ranking, and output writing. Post-hoc accuracy
evaluation and correctness audits are excluded, as are queue waiting and
prior PubChem download and catalog ingestion.
The MassSpecGym measurement is accumulated processing time of
\MsgPipelineSeconds{} seconds, or \MsgPipelinePerSpectrum{} seconds per
spectrum, and excludes startup and file I/O. Table~\ref{tab:pipeline_timing}
reports these runtimes alongside their component costs.

\subsection{Search Timing Boundaries}

Search timers start after the main-process model has been initialized and
include worker initialization, warm-up, and search until results are returned
in memory. NPLIB1 processes the full test set in one task. MassSpecGym is
divided into sixteen contiguous groups, with at most two tasks running
concurrently. Reported time is the sum of all sixteen task durations. Each
task uses 32 physical CPU cores, 32 workers, and 192 GiB RAM. Learned search
also uses one H100.

\subsection{Pipeline Timing Boundaries}
\label{app:pipeline_timing}

The NPLIB1 $P=B=16$ GEMS timer starts at process launch and stops after
ranked predictions have been written and the process exits. It measures one
continuous execution from available raw spectra and formula catalogs.
All three runs start fresh processes and reload the models; filesystem
caches are not flushed. The end-to-end value is used consistently in
Tables~\ref{tab:direct_search} and~\ref{tab:pipeline_timing}.

The component rows in Table~\ref{tab:pipeline_timing} isolate in-memory
processing costs. Preprocessing parses and normalizes spectrum peaks;
annotation assigns molecular subformulas; MIST predicts fingerprint
probabilities. Database construction freshly fingerprints validated
candidate SMILES, and retrieval selects the starting structures. Final
ranking includes consensus scoring, molecule validation, connectivity
deduplication, and worker startup and shutdown. Input-component and ranking
times are medians of three repetitions, as is the NPLIB1 $P=B=16$ search
time; other search times retain the recorded task measurements.

Continuous end-to-end timing is available only for NPLIB1 GEMS at $P=B=16$.
The remaining entries in Table~\ref{tab:direct_search} report accumulated
processing costs for the shared inputs, each method's search, and final
ranking, excluding startup, file I/O, and post-hoc evaluation. Thus the
NPLIB1 comparison charges loading and output overhead to GEMS alone.
MassSpecGym search costs cover sixteen tasks, with final ranking measured
separately over the full cohort on 32 CPU workers.

\begin{table}[htbp]
  \caption{Runtime measurements for $K=5$, $P=B=16$, and $D=3$, using one H100 and 32 physical CPU cores. NPLIB1 reports median end-to-end wall time, including startup, loading, and output writing. MassSpecGym reports accumulated processing time excluding this overhead; its search covers 16 tasks. Component rows show in-memory processing costs. Per-spectrum values divide the reported runtimes by 803 or 17,556 spectra.}
  \label{tab:pipeline_timing}
  \centering
  \scriptsize
  \setlength{\tabcolsep}{4pt}
  \begin{tabular}{lrr}
    \toprule
    Component & NPLIB1 (s) & MassSpecGym (s) \\
    \midrule
    Spectrum preprocessing & 0.57 & 0.40 \\
    Peak annotation & 0.61 & 3.04 \\
    MIST prediction & 0.40 & 4.91 \\
    Build fingerprint database from SMILES & 28.01 & 77.41 \\
    Retrieve top-5 seeds & 0.32 & 1.50 \\
    GEMS search & 129.28 & 3,869.90 \\
    Final ranking & 6.06 & 133.85 \\
    \midrule
    Reported runtime & 200.12 & 4,091.00 \\
    Per spectrum & 0.249 & 0.233 \\
    \bottomrule
  \end{tabular}
\end{table}
\FloatBarrier

\paragraph{FRIGID-base generation.}
A separate FRIGID-base generation run over all 803 NPLIB1 spectra
took \NplibFrigidBaseSeconds{} seconds on one H100. This standalone timing is distinct
from the run supplying the initial pool for the paired FRIGID/GEMS
refinement comparison. FRIGID's published accuracy in
Table~\ref{tab:main_results} includes subsequent ICEBERG refinement.

\paragraph{FRIGID with ICEBERG refinement.}
Our NPLIB1 run takes \NplibFrigidWallSeconds{} seconds, approximately
\NplibFrigidWallHours{} hours, for all 803 test spectra. It uses one
H100, 32 allocated CPU cores, and 128 GB host memory, with 128 attempted
samples per spectrum per round and initial
generation followed by 24 refinement rounds. The final saved ranking gives
Hit@1/Hit@10 of \NplibIcebergNativeHitOne\%/\NplibIcebergNativeHitTen\%.
This benchmark wall time includes model and data loading, MIST prediction,
generation, ICEBERG subprocesses, per-round ranking and evaluation, and
output writing. GEMS's end-to-end runtime is \NplibPipelineSeconds{} seconds,
approximately \NplibPipelineMinutes{} minutes, using one H100, 32 physical
CPU cores, and 192 GiB host memory. It includes loading and writing through
the final ranked predictions, with post-hoc evaluation excluded as described
above.

\paragraph{ICEBERG reranking.}
\label{app:iceberg_timing}
Producing the test prediction caches for Appendix~\ref{app:iceberg_reranking}
takes 16 minutes 58 seconds on one H100 for NPLIB1 and 9 minutes 59 seconds
across eight H100s for MassSpecGym, totaling 17.0 and 79.2 GPU-minutes,
respectively. These prediction-job times include model setup and output
writing, but exclude queue waiting, input preparation, GEMS search, and
coefficient selection. CPU spectral scoring adds approximately 15 and
81 seconds, respectively. The caches support multiple scoring and energy
choices. The selected rules require only 60,622 of 69,425 cached conditions
on NPLIB1 and 106,676 of 355,912 on MassSpecGym.

\subsection{Illustrative Neighborhood Enumeration}
\label{app:neighborhood_example}

For a 32-heavy-atom NPLIB1 molecule, one-edit enumeration produced 1,904
distinct connectivities in 2.10 seconds. Extending to two edits processed
4.57 million valid proposals and found 1.72 million additional connectivities
in 41 minutes 39 seconds on one CPU worker.\footnote{The
example is $\mathrm{C}_{24}\mathrm{H}_{30}\mathrm{F}_{2}\mathrm{O}_{6}$,
InChIKey14 \texttt{FEBLZLNTKCEFIT}.}
A third exhaustive round would expand 2.20 million atom-aligned states,
with an upper bound of 5.68 billion raw edit proposals before validity filtering.
This rapid neighborhood growth motivates GEMS's selective edit expansion
and bounded beam at each depth.

\FloatBarrier
\clearpage
\section{Illustrative Search Trajectories}
\label{app:trajectory_examples}

\begin{figure}[htbp]
  \centering
  \includegraphics[width=\linewidth]{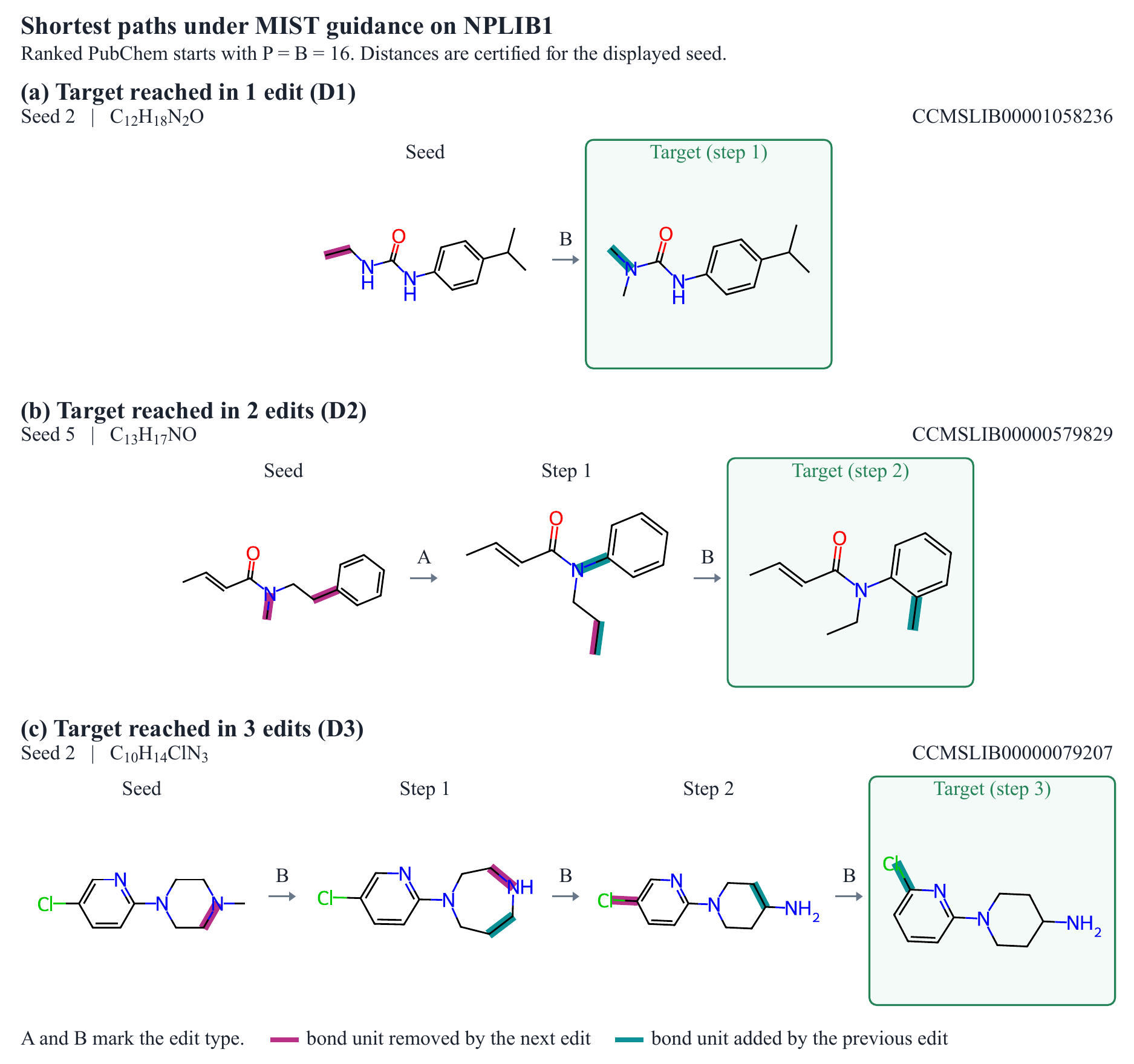}
  \caption{\textbf{NPLIB1 paths from seeds at certified distances D1--D3.}
  MIST-guided search reaches the target in one, two, and three edits,
  respectively, matching the minimum distance from each displayed seed.
  The examples show side-chain rearrangement, changes in substituent
  attachment, and ring expansion followed by contraction.
  All use the reported checkpoint, ranked PubChem starts, and $P=B=16$.
  A/B denote edit types. Magenta highlights mark bond-order decreases in
  the outgoing edit, and teal marks increases in the incoming edit.
  These are graph-edit paths, not proposed chemical reactions.}
  \label{fig:nplib_short_paths}
\end{figure}

\begin{figure}[htbp]
  \centering
  \includegraphics[width=\linewidth]{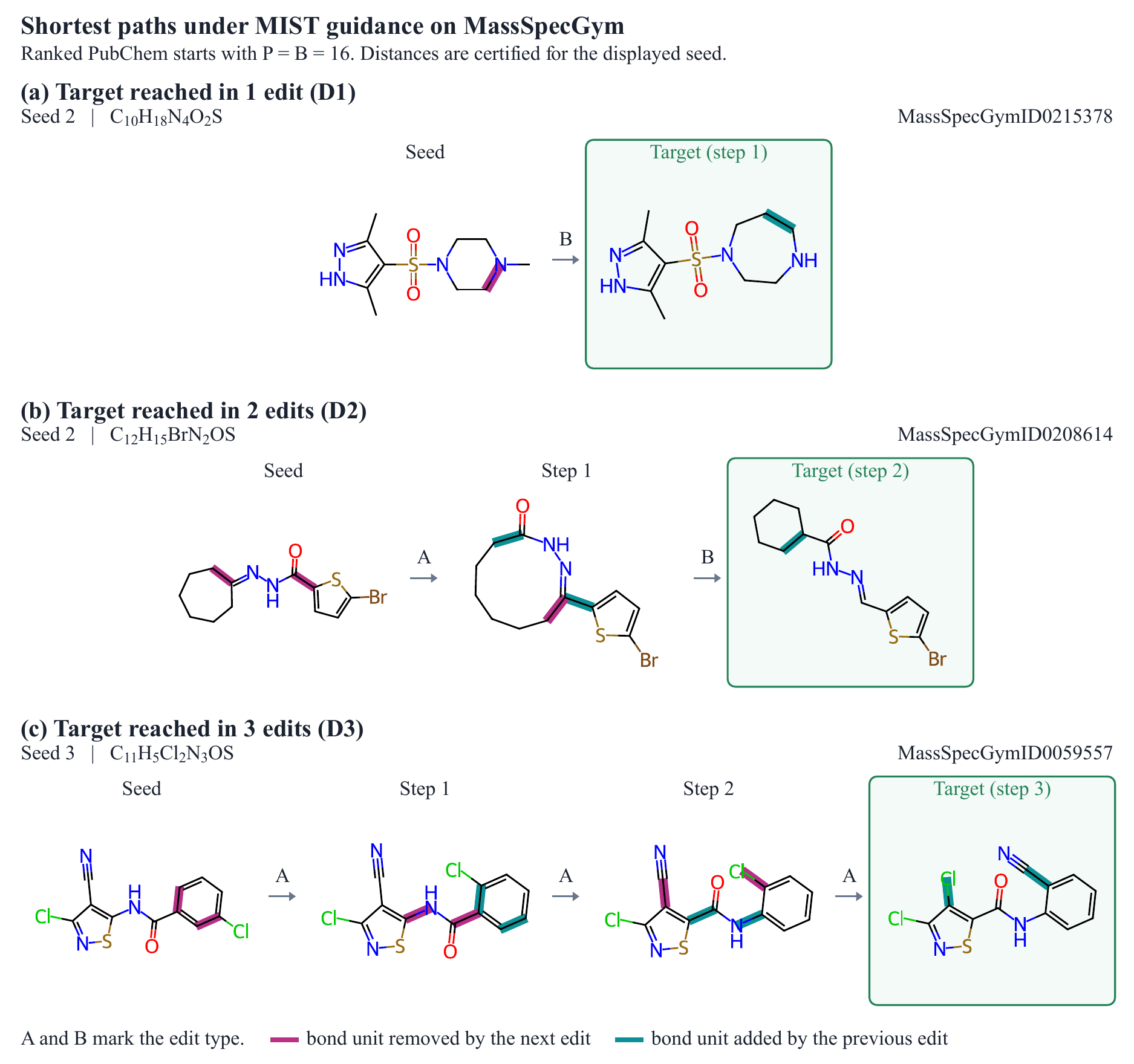}
  \caption{\textbf{MassSpecGym paths from seeds at certified distances D1--D3.}
  MIST-guided search follows a shortest path from each displayed seed.
  The examples include ring expansion, rearrangement through a larger-ring
  intermediate, and redistribution of substituents between aromatic rings.
  All use the reported checkpoint, ranked PubChem starts, and $P=B=16$.
  Display conventions follow Figure~\ref{fig:nplib_short_paths}.}
  \label{fig:msg_short_paths}
\end{figure}

Figures~\ref{fig:nplib_short_paths} and~\ref{fig:msg_short_paths} show
MIST-guided paths from selected PubChem seeds at certified distances one,
two, and three on each benchmark. Distance refers to the displayed seed,
not the closest seed in the query's starting pool. Each sequence follows a
retained seed lineage to its first target hit. All displayed edits were
replayed from the saved initial state, checking formula and atom-valence
preservation and the final target InChIKey14. The shorter-path exclusion
procedure is described in Appendix~\ref{app:certified_navigation}.
These examples were selected after evaluation for illustration.

\FloatBarrier

\FloatBarrier

\begin{figure}[htbp]
  \centering
  \includegraphics[width=\linewidth]{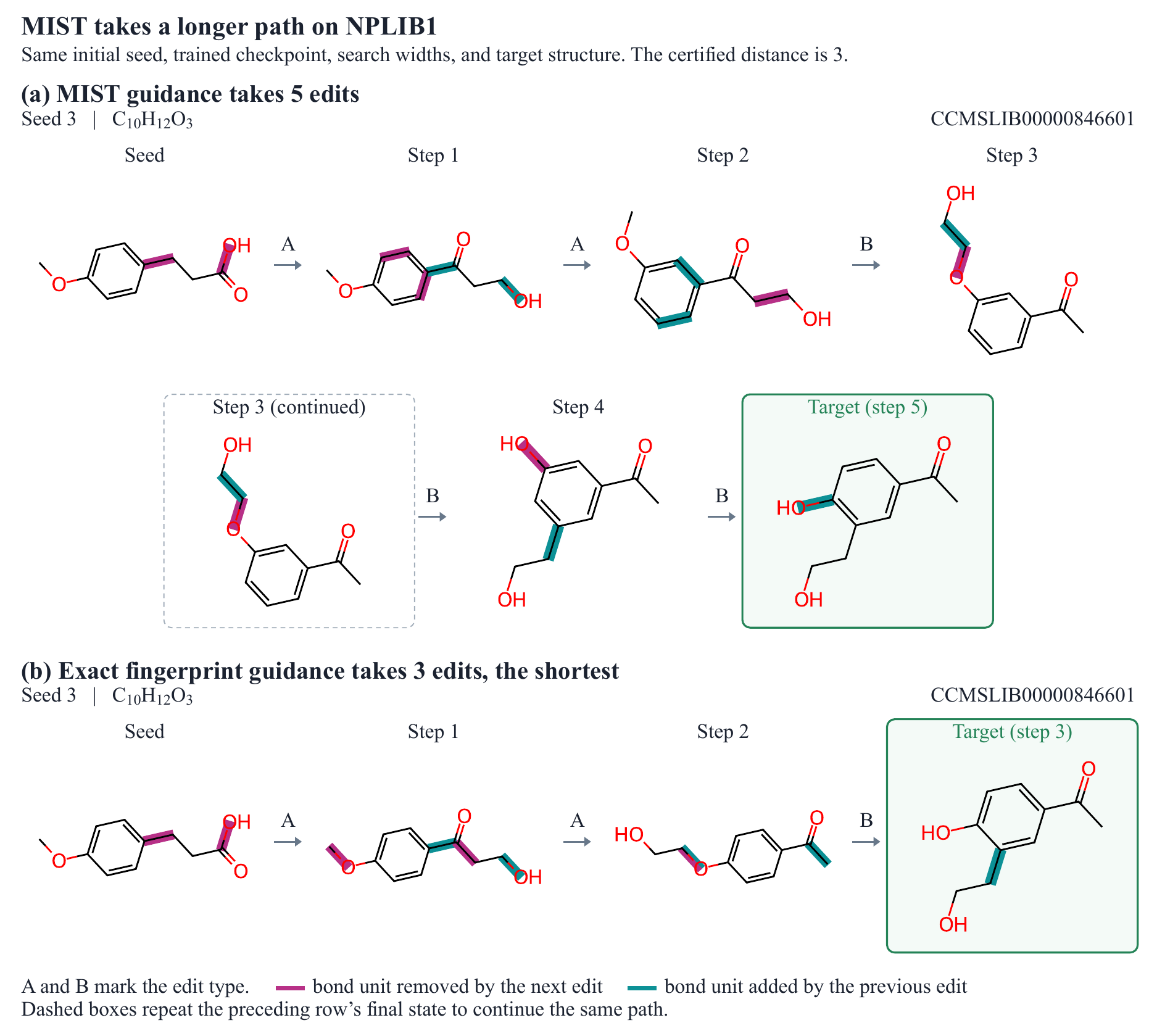}
  \caption{\textbf{A five-edit MIST path and a three-edit exact-guided path on NPLIB1.}
  Both paths start from Seed~3 of \texttt{CCMSLIB00000846601}.
  They share their first edit and then diverge. Exact guidance reaches the
  target in the certified minimum of three edits, while MIST first reaches
  it after five. Search uses $P=B=16$ with a ten-round budget.
  The displayed paths stop at the first hit from this seed.
  Dashed boxes repeat the preceding row's final state to continue the same path.}
  \label{fig:nplib_guidance_paths}
\end{figure}

\begin{figure}[htbp]
  \centering
  \includegraphics[width=\linewidth]{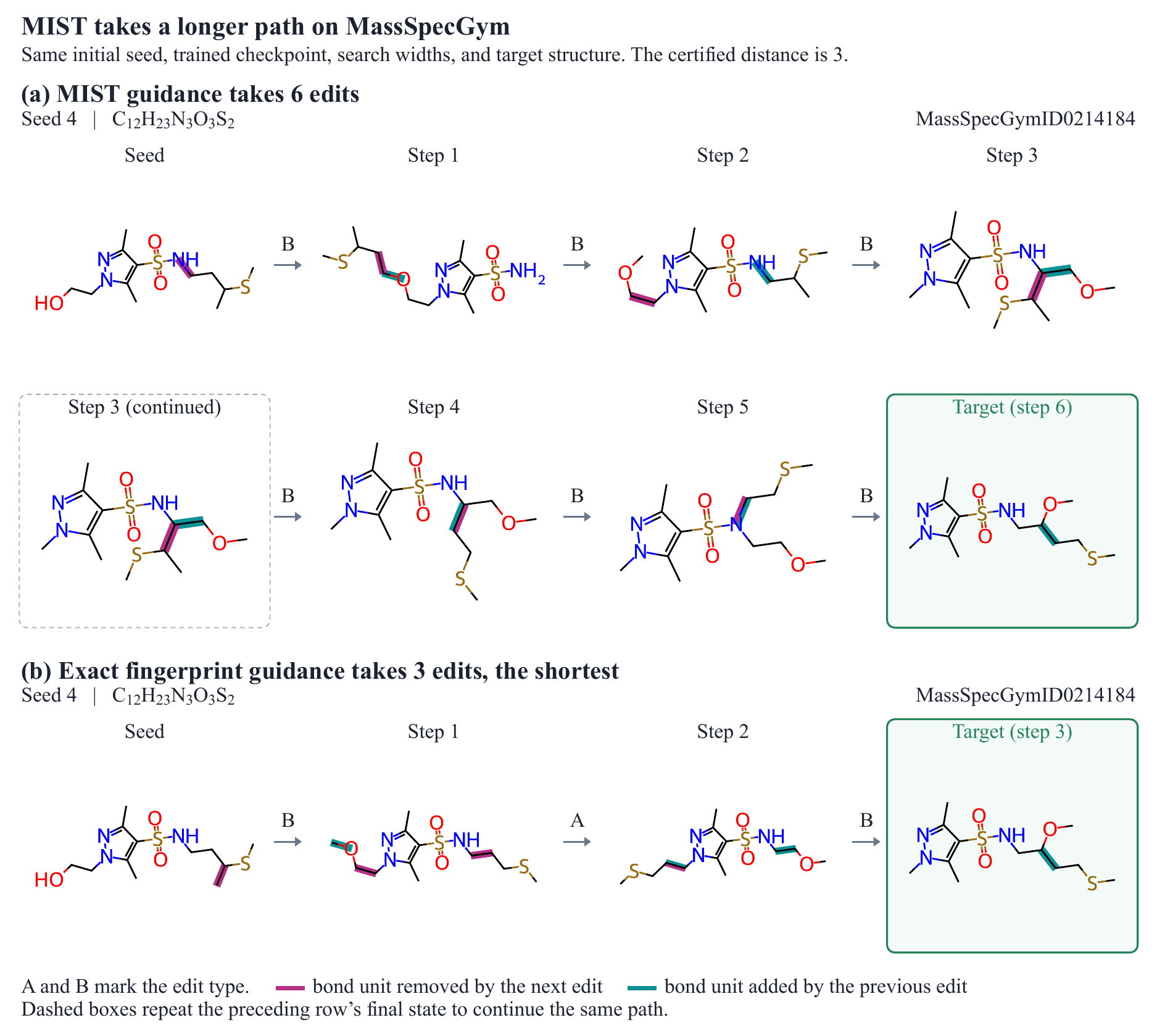}
  \caption{\textbf{A six-edit MIST path and a three-edit exact-guided path on MassSpecGym.}
  Both paths start from Seed~4 of \texttt{MassSpecGymID0214184} and
  end at the same target structure. Exact guidance follows a shortest path.
  MIST takes three additional edits in this search before its first retained
  target hit from this seed. Search uses $P=B=16$ with a ten-round budget.
  Display conventions follow Figure~\ref{fig:nplib_guidance_paths}.}
  \label{fig:msg_guidance_paths}
\end{figure}

\paragraph{Longer paths under predicted fingerprints.}
Figures~\ref{fig:nplib_guidance_paths} and~\ref{fig:msg_guidance_paths}
compare guidance from MIST predictions and exact target fingerprints.
Within each pair, the initial seed state, trained checkpoint, search widths,
and final target structure are identical. The first retained target path
from that seed takes five versus three edits on NPLIB1 and six versus three
on MassSpecGym. The certified distance is three in both cases, so the
MIST-guided paths contain two and three additional edits, respectively.
These comparisons concern navigation; they do not depend on the target's
final rank or imply that another seed could not reach it sooner.

\FloatBarrier

\FloatBarrier

\begin{figure}[htbp]
  \centering
  \includegraphics[width=\linewidth]{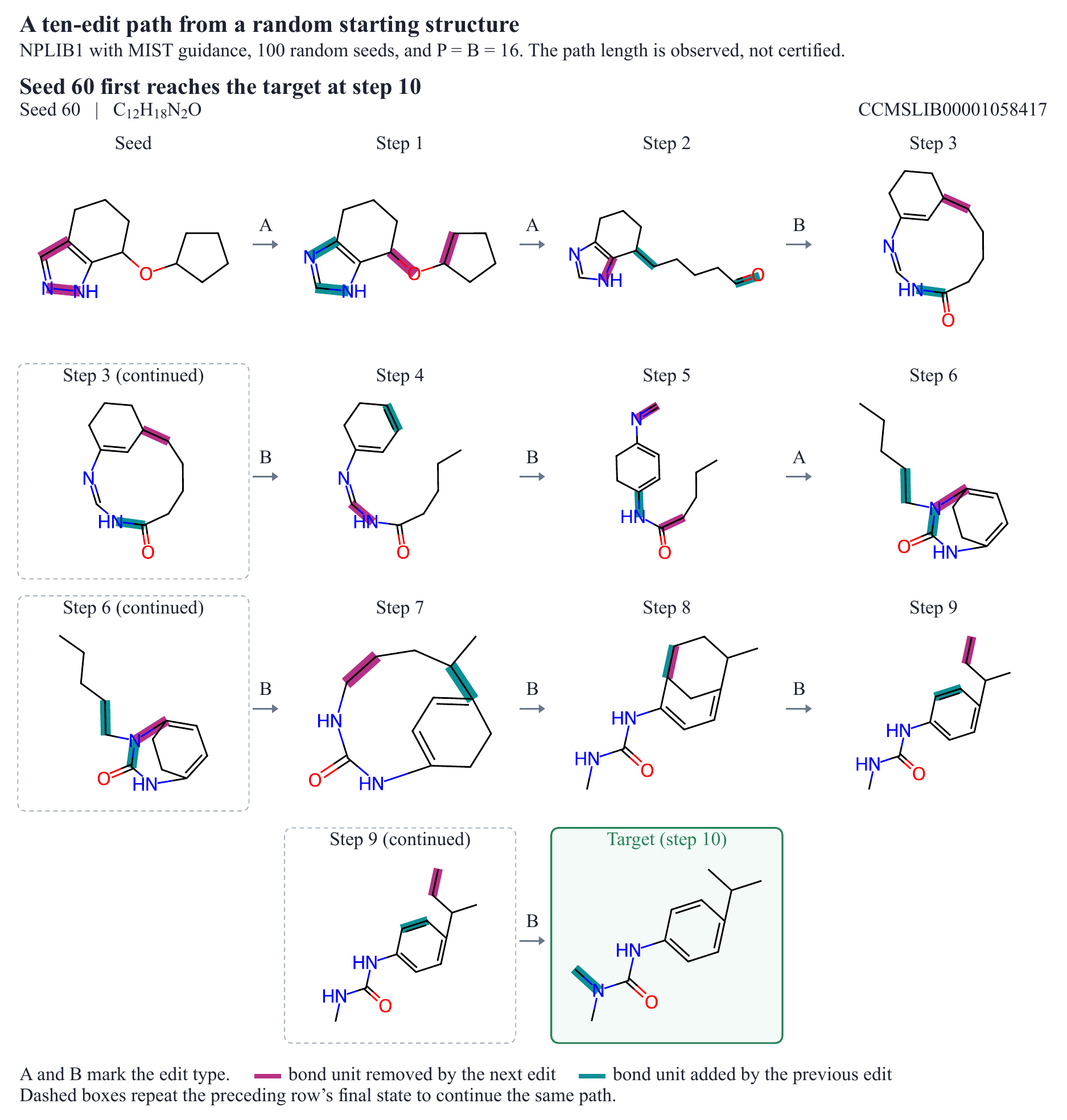}
  \caption{\textbf{A ten-edit MIST-guided path from a random seed.}
  Seed~60 of the 100-seed random pool first reaches the target at depth ten
  for NPLIB1 query \texttt{CCMSLIB00001058417}.
  Every displayed transition is a replayed retained edge from the
  $P=B=16$, $D=10$ search. The example illustrates substantial structural
  rearrangement along a successful lineage, rather than a certified
  ten-edit distance or a requirement for ten rounds across all seeds.}
  \label{fig:random100_long_path}
\end{figure}

\paragraph{A longer path from a random start.}
Figure~\ref{fig:random100_long_path} follows one of 100 random PubChem
starts for an NPLIB1 query. The retained path opens and reconnects rings
before reaching the target at its tenth edit. This is an observed path
length; its shortest possible distance has not been certified.
Other seeds for this query recover the target earlier.

\FloatBarrier

\end{document}